\documentclass{article} % For LaTeX2e

\PassOptionsToPackage{numbers, sort&compress}{natbib}
\usepackage[preprint]{neurips_2022}

\usepackage[utf8]{inputenc} % allow utf-8 input
\usepackage[T1]{fontenc}    % use 8-bit T1 fonts
\usepackage[pagebackref,breaklinks,colorlinks]{hyperref}       % hyperlinks
\usepackage{url}                        % simple URL typesetting
\usepackage{booktabs}                   % professional-quality tables
\usepackage{amsfonts}                   % blackboard math symbols
\usepackage{nicefrac}                   % compact symbols for 1/2, etc.
\usepackage{microtype}                  % microtypography
\usepackage[dvipsnames]{xcolor}         % colors
\usepackage{graphicx}
\usepackage{amsmath}
\usepackage{amssymb}

\usepackage[capitalize]{cleveref}
\crefname{section}{Sec.}{Secs.}
\Crefname{section}{Section}{Sections}
\Crefname{table}{Table}{Tables}
\crefname{table}{Tab.}{Tabs.}

\usepackage{subcaption}

\usepackage{xspace}
\makeatletter
\DeclareRobustCommand\onedot{\futurelet\@let@token\@onedot}
\def\@onedot{\ifx\@let@token.\else.\null\fi\xspace}

\def\eg{\emph{e.g}\onedot} 
\def\ie{\emph{i.e}\onedot} 
 
\def\etc{\emph{etc}\onedot} 
\def\wrt{w.r.t\onedot}

\makeatother

\usepackage[frozencache,cachedir=minted-cache]{minted}
\newminted{python}{frame=single,fontsize=\footnotesize,breaklines,autogobble,breakafter={(.},rulecolor=\color{black!40},style=emacs}
\newminted{yaml}{frame=single,fontsize=\footnotesize,breaklines,autogobble,breakanywhere,rulecolor=\color{black!40}}

\usepackage{enumitem}
\usepackage{mdframed}
\usepackage{listings}
\usepackage{multirow}
\usepackage{siunitx}
\def \numk #1k{\qty[round-mode=none]{#1}{\kilo{}}}
\def \numm #1m{\qty[round-mode=none]{#1}{\mega{}}}
\usepackage{pifont}% http://ctan.org/pkg/pifont
\newcommand{\cmark}{\ding{51}}%
\newcommand{\xmark}{\ding{55}}%

\lstnewenvironment{inst}
{\lstset{
  basicstyle=\ttfamily\footnotesize,
  tab=,
  tabsize=2,
  frame=single,
  backgroundcolor=\color{white},
  rulecolor=\color{black!40},
  alsoletter={[]:=<>},
  breaklines=true,
  breakindent=0pt,
  breakautoindent=false,
  moredelim=[s][\color{blue}\ttfamily\footnotesize]{[}{]},
  postbreak=\mbox{\textcolor{red}{$\hookrightarrow$}\space},
}}
{}
\lstnewenvironment{plaintext}
{\lstset{
  basicstyle=\ttfamily\footnotesize,
  tab=,
  tabsize=2,
  frame=single,
  backgroundcolor=\color{white},
  rulecolor=\color{black!40},
  alsoletter={:|},
  breaklines=true,
  breakindent=0pt,
  breakautoindent=false,
}}
{}
\newcounter{illus}
\renewcommand{\theillus}{\arabic{illus}}
\newenvironment{illus}[2][]{%
\refstepcounter{illus}
\ifstrempty{#1}%
{\mdfsetup{frametitle={\textit{Illustration~\theillus.}}}}% if condition (without title)
{\mdfsetup{frametitle={\textit{Illustration~\theillus.}~#1}}}% else condition (with title)
\mdfsetup{backgroundcolor=gray!20,linecolor=white,frametitlefont=\normalfont,frametitlebelowskip=0}
\begin{mdframed}[]\relax\label{#2}}{%
\end{mdframed}}

\Crefname{illus}{Illustration}{Illustrations}
\crefname{illus}{Illus.}{Illus.}
\def\ofasys{OFASys} % times doesn't provide slated small caps which looks weird in Abstract with style
\def\ofasysstyled{\texttt{\textsc{\ofasys}}}
\newcommand{\concept}[1]{``\textit{#1}''}
\newcommand{\param}[1]{\texttt{#1}}

\let\func\param
\newcommand{\tablecell}[2]{\begin{tabular}{@{}#1@{}}#2\end{tabular}}

\newcommand{\placeholder}[1]{\textcolor{blue}{[placeholder for #1]}}
\newcommand{\reviewcheck}[1]{\textcolor{red}{[check this ``#1'']}}
\newcommand{\zc}[1]{\textcolor{red}{[zc comments ``#1'']}}

\newcommand{\eat}[1]{}

\begin{document}

% \title{\ofasys: Scale beyond Modalities and Tasks}
\title{
%\ofasysstyled: An Open-Source Library for \\ Unified Multi-Modal Multi-Task Learning \\ 
\ofasysstyled: A Multi-Modal Multi-Task Learning System for Building Generalist Models}

\author{Jinze Bai\thanks{Equal contribution}, Rui Men\footnotemark[\value{footnote}],  Hao Yang\footnotemark[\value{footnote}], 
Xuancheng Ren\footnotemark[\value{footnote}], Kai Dang, Yichang Zhang,\\ 
\bf Xiaohuan Zhou, Peng Wang, Sinan Tan, An Yang, Zeyu Cui, Yu Han, Shuai Bai, \\
\bf Wenbin Ge, Jianxin Ma, Junyang Lin, Jingren Zhou, Chang Zhou\thanks{Corresponding Author} \\
Damo Academy, Alibaba Inc. \\ % / {\tt \{first-name\}@alibaba-inc.com}\\
% First line of institution1 address\\
\tt\small \{jinze.bjz, menrui.mr, yh351016, renxuancheng.rxc, dangkai.dk, \\
\tt\small yichang.zyc, shiyi.zxh, zheluo.wp, tansinan.tsn, ya235025, zeyu.czy, \\ 
\tt\small hanyu.han, baishuai.bs, gewenbin.gwb, jason.mjx, junyang.ljy, \\
\tt\small jingren.zhou, ericzhou.zc\}@alibaba-inc.com
% \and
% OFASys Authors as well\\
% Damo Academy, Alibaba Inc.\\
% % First line of institution2 address\\
% {\tt\small secondauthor@alibaba-inc.com}
}
\maketitle

\begin{abstract}
Generalist models, which are capable of performing diverse multi-modal tasks in a task-agnostic way within a single model, have been explored recently.
Being, hopefully, an alternative to approaching general-purpose AI, existing generalist models are still at an early stage, where modality and task coverage is limited. %while the task relationships are also under-explored. 
To empower multi-modal task-scaling and speed up this line of research, we release a generalist model learning system, \ofasysstyled{}, built on top of a declarative task interface named \textit{multi-modal instruction}. 
At the core of \ofasysstyled{} is the idea of decoupling multi-modal task representations from the underlying model implementations.
In \ofasysstyled{}, a task involving multiple modalities can be defined declaratively even with just a single line of code. 
The system automatically generates task plans from such instructions for training and inference. 
It also facilitates multi-task training for diverse multi-modal workloads. 
As a starting point, we provide presets of 7 different modalities and 23 highly-diverse example tasks in \ofasysstyled{}, with which we also develop a first-in-kind, \textit{single} model, OFA+, that can handle text, image, speech, video, and motion data. 
The single OFA+ model achieves 95\% performance in average with only 16\% parameters of 15 task-finetuned models, 
%showcasing the potential of multi-modal task-scaling with \ofasysstyled{}.
showcasing the performance reliability of multi-modal task-scaling provided by \ofasysstyled{}.
%By using a simple multi-task schedule, OFA+ achieves 94\% of the performance in average with only 5\% parameters of the task-finetuned models, further showcasing the potential of generalist models.

\smallskip
\centering
Available at \url{https://github.com/OFA-Sys/OFASys}.
\end{abstract}
%%%%%%%%% BODY TEXT

\section{Introduction}
\label{sec:intro}

% Deep learning researches have achieved great success in pursuing ultimate performance on a narrow field of tasks~\cite{srivastava22bigbench}, by designing curated model structures, data representations and training objectives for each task. 
Deep learning researches have achieved great success in designing curated model structures, data representations, and training objectives to pursue ultimate performance for a single model on a narrow field of tasks~\cite{srivastava22bigbench}. 
Recently, generalist models, e.g., OFA~\cite{wang22ofa}, Flamingo~\cite{alayrac2022flamingo}, GATO~\cite{reed2022generalist} and Unified-IO~\cite{lu2022unifiedio}, have been working towards a different vision of performing diverse multi-modal tasks in a task-agnostic way within a single model. 

Inspired by the success of large language models~\cite{brown20gpt3,raffel20t5,chowdhery22palm}, 
%and their broad applicability to vision~\cite{beit,mae}, audio~\cite{hsu2021hubert,wav2vec2.0} and multi-modal~\cite{wang22beitv3} domains
a generalist model expresses any task intention via natural language and provides unified representations for the same modality data across all tasks. 
Such general task representation and task-agnostic learning, which is considered the prerequisites to approaching general-purpose AI~\cite{minsky1988society}, have inspired
the potential of generalizing to unseen tasks even with different modality compositions. 
%Finding such general representations is considered the prerequisites to approaching general-purpose AI~\cite{minsky1988society}, 
Despite the disadvantage that no task-specific model structures are introduced in the finetuning stage, these generalist models have demonstrated unprecedentedly the possibilities of achieving superior finetuning performance, by task-agnostic multi-modal and multi-task pretraining.
\begin{table}[t]
    \centering
    % \small
    \begin{tabular}{@{} l c c c c @{}}
    \toprule
      Model    & Paradigm & Supervised & Multi-Task & Multi-Modal \\\midrule
      GPT~\cite{radford18gpt,radford19gpt2,brown20gpt3}      & CLM       & \xmark     & \xmark     & \xmark  \\
      PaLM~\cite{chowdhery22palm} & CLM & \xmark & \xmark & \xmark \\
      FLAN~\cite{wei22finetuned}    & CLM       & \cmark     & \cmark     & \xmark  \\
      Flamingo~\cite{alayrac2022flamingo} & CLM       & \xmark     & \cmark     & \phantom{(0)} \cmark (3)  \\
      Gato~\cite{reed2022generalist}     & CLM       & \cmark     & \cmark     & \phantom{(0)} \cmark (4)\\
      T0~\cite{sanh22t0}       & S2S      & \cmark     & \cmark     & \xmark  \\
      OFA~\cite{wang22ofa}      & S2S      & \cmark     & \cmark     & \phantom{(0)} \cmark (3)  \\
      Unified-IO~\cite{lu2022unifiedio} & S2S    & \cmark     & \cmark     & \phantom{(0)} \cmark (4)  \\ \midrule
      \ofasysstyled{} & CLM, S2S, DDPM & \cmark & \cmark & \phantom{(0)} \cmark (7) \\
      \bottomrule
    \end{tabular}
    
    \caption{Approaches to building generalist models. \ofasysstyled{} covers all of the approaches. CLM stands for causal language modeling, S2S stands for sequence-to-sequence learning, and DDPM stands for denoising diffusion probabilistic modeling. The numbers in parentheses are the numbers of supported modalities. The unique modalities supported by \ofasysstyled{} are \texttt{AUDIO} and \texttt{MOTION}.}
    \label{tab:comparison}
\end{table}
%%%%%%%%%%%%%%%%

%However, these generalist models are still highly limited in task population, task diversity and modality coverage which we argue might be the key to approach such vision. 
However, being at an early stage, the research on generalist models is much more complicated in engineering compared to that on task-specific models, since it requires systematic management of the relationships among multiple modalities and tasks, in addition to the neural architectures and the single task optimization. This engineering complexity becomes intractable as the modality and task population grow to larger scale.
While both open-source tensor libraries, such as TensorFlow~\cite{abadi2016tensorflow} and PyTorch~\cite{paszke2019pytorch}, and domain-specific libraries, such as Transformers~\cite{wolf2020transformers} and MMDetection~\cite{chen2019mmdetection}, greatly expedite the development of task-specific models and applications, there is currently no designated system that provides neat abstractions and tools for task-agnostic generalist model learning.

%Although existing libraries may be adapted to conduct these studies, it is often inconvenient and cumbersome with a library designed for a single task with a fixed number of modalities in mind.

The inherent diversity and heterogeneity of multi-modal multi-task learning stands out as a major obstacle in establishing such a system.
% The lack of such a system may result from the inherent diversity and heterogeneity of multi-modal multi-task learning.
% Nonetheless, the success of the existing generalist models justifies the feasibility of developing such a system or library, as each model develops its own system to manage the differences among tasks and modalities.
%On the one hand, the unified task formulation proves helpful to mitigate the task differences, especially the semantic task descriptions~\cite{sanh22t0,wei22finetuned,wang22ofa}.
%On the other hand, the unified data formulation (representation-based~\cite{wang22ofa,reed2022generalist} or discrete code--based~\cite{reed2022generalist,lu2022unifiedio,wang22ofa}) proves viable to incorporate multi-modal data to a single model.
In conventional practice, different tasks may require a different model structure and a different training pipeline, all contributing to the state-of-the-arts performance per task.
% Conventional machine learning systems are task specific in that a different kind of tasks requires a different kind of system designs.
Although the particular traits of the specific task can be efficiently addressed, it proves hard to scale, as each new task would demand a new system design.
%The incompatibility severely limits the possibility of multi-task learning and pose a prominent challenge for the system design.

Recently, studies have shed light on that issue and provided an alternative, scalable way to multi-modal multi-task learning, drawing inspirations from (a) the task generalization capabilities demonstrated by pretrained language models~\cite{brown20gpt3,sanh22t0} and (b) the success of the Transformer architecture in universal multi-modal learning~\cite{wang22ofa,wang22beitv3,radford21clip,chen2022pix2seqv2}.
This motivates us to decouple the task representation from its model implementation, which enables researchers to investigate multi-modal task scaling and the underlying model compositions independently. 
In light of this, we propose \ofasysstyled, a system designed for building generalist models via multi-modal multi-task learning. 
The goal of \ofasysstyled{} is to facilitate the research of multi-modal multi-task learning with a concise, flexible user interface and a modularized, reusable system design.

% Based on the unified design presented in OFA, \ofasysstyled{} supports training, evaluating, and inference in multi-modal multi-task learning, with even more modalities and tasks.
% \ofasysstyled{} currently includes the presets for \reviewcheck{22} tasks in \reviewcheck{7} modalities, which can trained all at once and \reviewcheck{is by far the most comprehensive open-source library}. 
% An overview of the library is shown in \cref{fig:overview}, which is elaborated in \cref{sec:design}.

% Striving to become a full-fledged library, \ofasysstyled{} supports training, evaluating, and inference in the multi-modal multi-task setup  and offers configuration-oriented, coding-free experience as well as easy-to-use programming interfaces. \ofasysstyled{} currently includes the presets for \reviewcheck{22} tasks in \reviewcheck{7} modalities, which can trained all at once and \reviewcheck{is by far the most comprehensive open-source library}.

\begin{figure}
\small
\centering

%%%%%%%% Text
\begin{subfigure}[b]{0.26\linewidth}
\textbf{Instruction (\texttt{TEXT} Only)}
\begin{inst}
what is the summary of article "[TEXT:src]"? -> [TEXT:tgt]
\end{inst}

\textbf{Data} \\
\textit{poland 's main opposition party tuesday endorsed president lech walesa in an upcoming presidential run-off election after a reformed communist won the first round of voting .}

\textbf{Result}  \\ 
polish opposition endorses walesa in presidential run-off
\caption{Text Summarization}
\end{subfigure}
\hfill
%%%%%%%%
\begin{subfigure}[b]{0.42\linewidth}
\textbf{Instruction (\texttt{IMAGE} \& \texttt{TEXT})}
\begin{inst}
what is the complete image? caption: "[TEXT:text]"? -> [IMAGE,preprocess=image_vqgan,adaptor=image_vqgan]
\end{inst} 

\textbf{Data} \\
a city with tall buildings and a large green park.

\textbf{Result}
\par\smallskip
\hfil
\includegraphics[height=8\baselineskip]{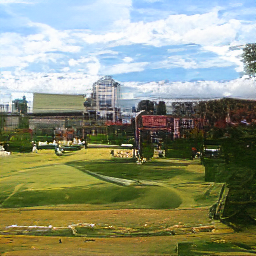} 
\hfil
\caption{Image Generation}
\end{subfigure}
%%%%%%%%
\hfill
%%%%%%%% image text bbox
\begin{subfigure}[b]{0.26\linewidth}
\textbf{Instruction (\texttt{IMAGE}, \texttt{TEXT} \& \texttt{BOX})}
\begin{inst}
[IMAGE:img] which region does the text "[TEXT:cap]" describe? -> [BOX:box]
\end{inst} 

\textbf{Data \& Result} \\
\texttt{cap:} \textit{taxi} 
\par\smallskip
\hfil
\includegraphics[trim={0 0 0 3.1cm},clip,width=0.93\linewidth]{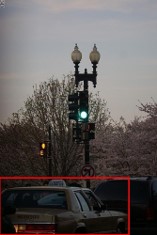} 
\hfil
\caption{Image Grounding}
\end{subfigure}
%%%%%%%%

\vspace{2\baselineskip}

\hfil
%%%%%%%% Video
\begin{subfigure}[b]{0.45\linewidth}
\textbf{Instruction (\texttt{VIDEO} \& \texttt{TEXT})}
\begin{inst}
[VIDEO:video] what does the video describe? -> [TEXT:cap]
\end{inst} 

\textbf{Data (Illustration)} \par\smallskip
\hfil
\includegraphics[width=.32\linewidth]{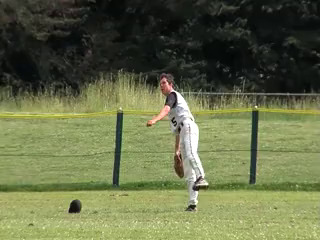}
\hfil
\includegraphics[width=.32\linewidth]{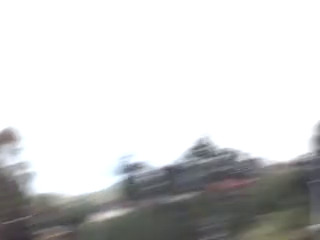}
\hfil
\includegraphics[width=.32\linewidth]{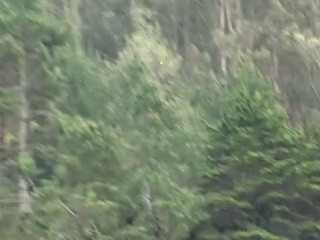}
\hfil

\hfil
\includegraphics[width=.32\linewidth]{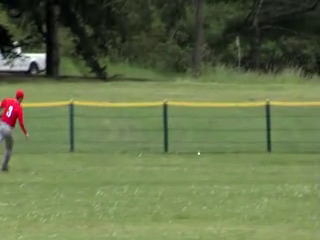}
\hfil
\includegraphics[width=.32\linewidth]{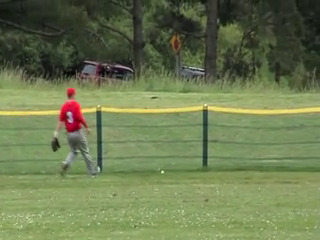}
\hfil
\includegraphics[width=.32\linewidth]{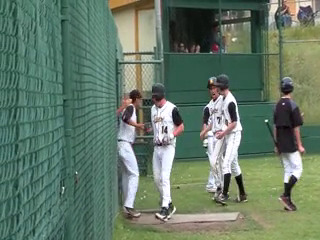}
\hfil

\textbf{Result} \\
a baseball player hits the ball

\caption{Video Captioning}
\end{subfigure}
%%%%%%%%
\hfil\hfil\hfil
%%%%%%%% Struct
\begin{subfigure}[b]{0.45\linewidth}
\textbf{Instruction (\texttt{STRUCT} \& \texttt{TEXT})}
\begin{inst}
structured knowledge: "[STRUCT:database]". how to describe the tripleset ? -> [TEXT:tgt]
\end{inst} 

\textbf{Data}

\vspace{.2\baselineskip}
% \resizebox{\linewidth}{!}{
{
\hfil
\scriptsize
\begin{tabular}{@{}lll@{}}
\toprule
Subject & Attribute       & Value    \\  
\midrule
Cocum   & eatType         & restaurant \\
Cocum   & food            & English   \\
Cocum   & priceRange      & moderate   \\
Cocum   & customer rating & \num{1} out of \num{5} \\
Cocum   & familyFriendly  & no \\
\bottomrule
\end{tabular}
% \begin{tabular}{@{}l@{}}
% \toprule
% Schema   \\  
% \midrule
% singer\_id, name, birth\_year, net\_worth\_millions, citizenship \\
% \bottomrule
% \end{tabular}
\hfil
}
% }
\vspace{.1\baselineskip}
% \relax\raggedright
% \textit{(Atlanta, OFFICIAL\_POPULATION, 5,457,831) \\
% ([TABLECONTEXT], METROPOLITAN\_\linebreak[0]{}AREA, Atlanta) \\
% (5,457,831, YEAR,  2012) \\
% ([TABLECONTEXT], [TITLE], List of metropolitan areas by population) \\
% (Atlanta, COUNTRY, United States)}

\textbf{Result} \\
% atlanta, united states has a population of 5,457,831 in 2012.
Cocum is an English restaurant with a moderate price range and a customer rating of \num{1} out of \num{5}. it is not family friendly.
\caption{Table-to-Text}
\end{subfigure}
%%%%%%%%
\hfil

\vspace{2\baselineskip}

\hfil
%%%%%%%% Audio
\begin{subfigure}[b]{.45\linewidth}
\textbf{Instruction (\texttt{AUDIO} \& \texttt{TEXT})}
\begin{inst}
[AUDIO:wav] what is the text corresponding to the voice? -> [TEXT:text,preprocess=text_phone]
\end{inst} 

\textbf{Data (Illustration)} 
\par\smallskip
\hfil
\includegraphics[width=.99\linewidth]{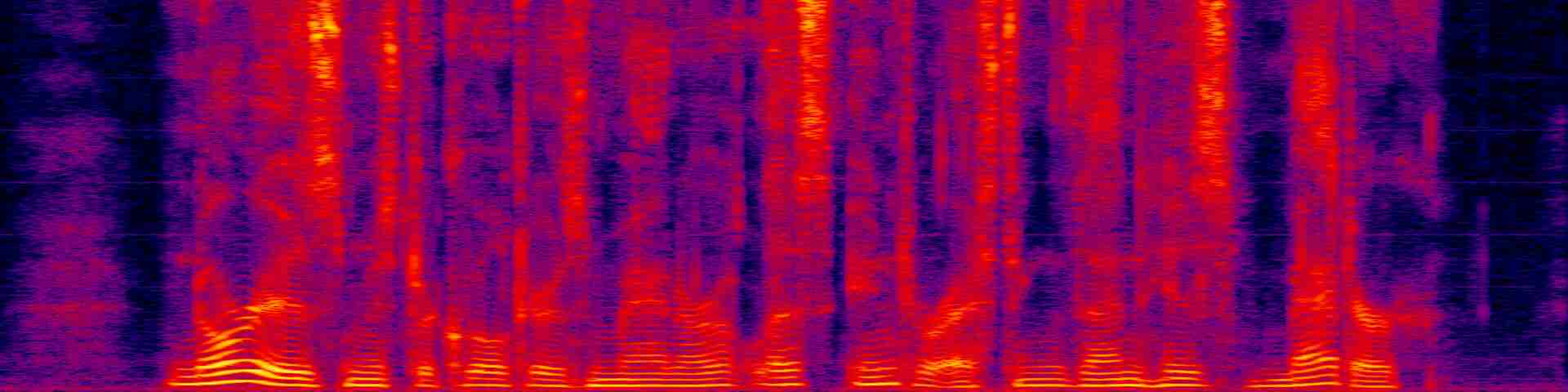}
\hfil

\textbf{Result} \\
nor is mister klohs manner less interesting than his manner
\caption{Automatic Speech Recognition}
\end{subfigure}
%%%%%%%%
\hfil\hfil\hfil
%%%%%%%% Motion
\begin{subfigure}[b]{.45\linewidth}

\textbf{Instruction (\texttt{MOTION} \& \texttt{TEXT})}
\begin{inst}
motion capture: [TEXT:text] -> [MOTION:bvh_frames]
\end{inst} 

\textbf{Data} \\
\textit{run and stop}

\textbf{Result (Illustration)} 
\par\medskip
\hfil
\includegraphics[width=.8\linewidth]{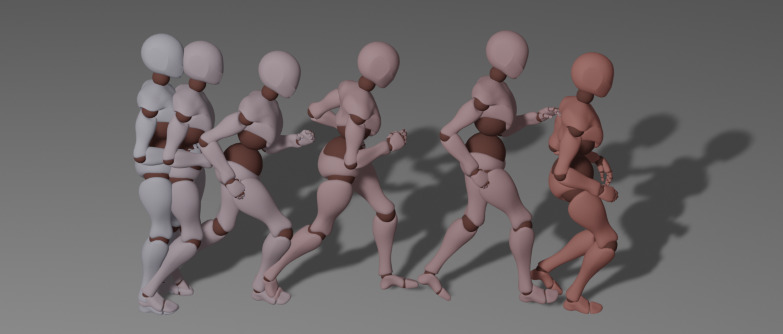}
\hfil
\caption{Text-to-Motion Synthesis}
\end{subfigure}
%%%%%%%%%%%%%%%
\hfil

\caption{\ofasysstyled{} enables extremely diverse compositions of modalities for tasks via the instruction interface. Results are generated by models trained by \ofasysstyled{}.}
\label{fig:overview-tasks-and-modalites}
\end{figure}

For users, \ofasysstyled{} enables both fast prototyping and in-depth customization via a declarative interface called \concept{Multi-Modal Instruction}.
The instruction describes a task using natural language with multi-modal data placeholders called \concept{slots}.
Users can declare a new task in just a single line of code, or customize task-specific processing and new modalities, which is seamlessly combined with the instruction interface.
%For basic usage, users can define a new task in just a single line of code.
%For a simple example, image captioning can be formulated as 
%``What is the caption of this image? [IMG] -\textgreater{} [TEXT]''
%(please refer to \cref{sec:usage} for a brief introduction).
%For advanced usage, users can easily introduce task-specific processing and new modalities programmatically, which is seamlessly combined with the instruction interface.
As a starting point, \ofasysstyled{} comes with \num{7} modality presets, \ie, \texttt{TEXT}, \texttt{IMAGE}, \texttt{AUDIO}, \texttt{VIDEO}, \texttt{STRUCT}, and \texttt{MOTION}, which compose \num{23} example tasks that vary widely in modality compositions and task objectives.
A brief demonstration of the available modalities and representative example tasks is shown in \cref{fig:overview-tasks-and-modalites}.
% In addition, \ofasysstyled{} provides a simple programming interface based on \textit{instructions}.
% Instructions significantly reduce the difficulty in defining multi-modal tasks for beginners.
% Instructions also support fine-grained control of modality processing via slot attributes for advanced users.
% \ofasysstyled{} is task-scalable in that once a processing pipeline for a modality is implemented, it can be automatically used for all tasks via slots.

%%%%%%%%%%%%%%%%
%%%%%%%%%%%%%%%%

% %%%%%%%% Image
% \begin{subfigure}[b]{0.3\linewidth}
% \textbf{Instruction (\texttt{IMAGE} \& \texttt{TEXT})}
% \begin{inst}
% [IMAGE:img] what does the image describe? -> [TEXT:cap]
% \end{inst} 

% \textbf{Data} 

% \par\smallskip
% \hfil
% \includegraphics[width=0.82\linewidth]{pictures/ex-captioning.jpg} 
% \hfil

% \textbf{Result}  \\ 
% a hand is holding an open book
% \caption{Image Captioning}
% \end{subfigure}
% %%%%%%%%

To realize the goal of \ofasysstyled{}, the system design, which disentangles the complexity of implementing task-specific pipelines, %consists of a hierarchy of reusable components:
forms reusable component hierarchies in different granularity:
(a) for different tasks, the model and the training/inference components can form training/inference pipelines, 
(b) for a single task, the universal model and the modality-specific model components can form a multi-modal model computing pipeline, and
(c) for a slot in a task, the pre-/post-processors and the adapters can form modality-specific data pipelines.
In addition, the multi-task engine provides support to efficient multi-task training.
% isolates the complexity of task-specific pipelines into three composable, reusable components:
% (a) the declarative task representation interface, where the task desideratum is expressed in natural language containing data placeholders for multi-modal data,
% (b) the universal compute engine, which is a representation-based neural network that can be either modality-agnostic or modality-aware, 
% with several training objectives and inference generators, and
% (c) the modality-specific IO, which maps the multi-modal data to and from representations and deals with various complication in data processing.

\ofasysstyled{} enables us to train a series of specialist and generalist models, which we name OFA+.
The OFA+ (Specialist) models showcase the capability of the instruction interface, with which one can expediently compose new tasks.
The OFA+ (Generalist) model spans over \num{7} modalities and is trained on 17 tasks, suggesting that training a single model with more modalities and tasks is achievable.
Moreover, the OFA+ (Generalist MoE) models with a sparsely-activated universal model demonstrates promising results.
By using a simple multi-task schedule, OFA+ (Generalist MoE) achieves \qty{95}{\percent} of the performance with just \qty{16}{\percent} of the parameters of task-finetuned models.

% an unprecedented generalist model, OFA+, spanning over \num{7} modalities, which is substantially diverse compared to existing generalist models. 
% The experimental results indicate that training a model with more modalities and more tasks is achievable and promising.
% By using a simple multi-task schedule, OFA+ achieves \qty{94}{\percent} of the performance with just \qty{6}{\percent} of the parameters of task-finetuned models

% The core contribution enabling the aforementioned capabilities is a declarative instruction interface that unifies and parameterized the processing of tasks and modalities at the library level.
% \textit{Instruction} consists of a template that describes the productive or operative task desiderata in natural language with placeholders called \textit{slots} representing inputs and outputs and related modality-specific processing. 
% For a simple example, the task of image description generation or image captioning can be formulated as ``What is the caption of this image? [IMG] -\textgreater [TEXT]''(the grammar of which is introduced in \cref{sec:instruction}).
% The abstraction allows \ofasysstyled{} to disentangle the task logic into a common backbone and modality-related customizable components.

The contribution of \ofasysstyled{} is summarized as follows:
\begin{itemize}
    \item \ofasysstyled{} is an open-source generalist model learning system that is designed for multi-modal task-scaling. \ofasysstyled{} offers to the community \num{7} pre-defined modalities and \num{23} example tasks for reuse in multi-modal multi-task learning research.
    \item \ofasysstyled{} provides an easy-to-use declarative interface that decouples task definitions from task implementations, thus enabling fast prototyping. The design of \ofasysstyled{} disentangles the complexity of single task pipelines into a hierarchy of components, which can be easily reused, customized, and replaced for in-depth research.
    \item \ofasysstyled{} trains the first-in-kind models named OFA+ that can understand and generate text, image, speech, video, and motion data. 
    The checkpoints are publicly available upon request. % at \todo{a url}. 
\end{itemize}

% The manuscript is organized as follows: \cref{sec:usage} introduces the basic usage of \ofasysstyled{} as a library; 
% \cref{sec:instruction} elaborates on the multi-modal instruction interface; \cref{sec:design} explains the library design and structure; \cref{sec:application} demonstrates the results obtained using \ofasysstyled{}; \cref{sec:discussion} discusses the scope and the current state of the library; and finally, \cref{sec:conclusion} concludes the paper. 
% For \cref{sec:usage,sec:instruction,sec:design,sec:application}, more details can be found in the appendix.

% \section{OFA in Review}

%\section{Usage Guide}
\section{Usage with Declarative Multi-Modal Instruction}
\label{sec:usage}
% \subsection{Multi-Modal Instruction}

%In the following, 
% We first introduce the overall usage of \ofasysstyled{} for general users (\cref{sec:usage-overall}) and then elaborate on the details \wrt the declarative instruction interface (\cref{sec:instruction,sec:training,sec:inference}).

% \subsection{High-Level Usage Overview}

% \label{sec:usage-overall}
%\ofasysstyled{} provides a unified task representation interface called \textit{multimodal instruction}. 
%\ofasysstyled{} aims to facilitate the experimentation of multi-modal multi-task learning with an easy-to-use user interface.
In the following, we briefly show the basic high-level usage of \ofasysstyled{} about how to represent, train, and conduct inference on multi-modal tasks with the
\textit{multi-modal instruction} interface. 
For more usage illustrations, please refer to \cref{sec:training,sec:inference,sec:extend}.
A multi-modal instruction is a descriptive line of code that specifies what the task is supposed to do and what kinds of modality of data are involved.
%The task description follows the sequence-to-sequence normative and the task data specification is defined in placeholders called \concept{slots}.
%Slot specifies the modality of the data and how should the data be processed for the model.
%A basic example is shown in \cref{illus:basic}.
%while the full instruction grammar and concept illustrations can be found at \cref{sec:instruction}.
%%%%%%%%
\begin{pythoncode}
instruction = "[IMAGE:img] what does the image describe? -> [TEXT:cap]"
\end{pythoncode}

\eat{
\begin{illus}[Basic Instruction for Image Captioning]{illus:basic}

\begin{inst}
[IMAGE:img] what does the image describe? -> [TEXT:cap]
\end{inst}

\noindent
The two sentences separated by \concept{-\textgreater} describe the task input and the task output, respectively. 
%The sentence before is for the encoder and the sentence after, the decoder.
%An instruction normally consists of two sentences separated by \concept{-\textgreater}. The sentence before is for the encoder and the sentence after, the decoder.
% A slot is a placeholder identified by `[]', which illustrates the meta-information of its modality inside. 
A slot, identified by ``[]'', specifies the modality and its corresponding column name. 
In this case, ``[IMAGE:img]'' specifies that there is an image input which binds to a data column named `img' in the dataset. 
The plain texts in the instruction indicate the task is about captioning an image.
The output of the task is a text sequence, which is the `cap' column in the dataset.
\end{illus}
}
%%%%%%%%

%\ofasysstyled{} begins with supports of \num{7} modalities/slot types, upon which we provide \num{23} example tasks across these modality, please refer to \cref{appx:slots} and \cref{appx:tasks}, respectively.

With the help of instructions, we can create multi-modal tasks as follows, for which \ofasysstyled{} determines the model structure and training/inference--related components automatically according to the preset choices or the given customized implementations:

%With the help of instructions, we can create each multi-modal task rapidly in one line of code, which is critical for multi-modal task scaling:
\begin{pythoncode}
from ofasys import Task, GeneralistModel, Trainer
task1 = Task(instruction1)
task2 = Task(instruction2)
\end{pythoncode}

The preceding tasks can then be bound to some dataset and join the training of a generalist model as
\begin{pythoncode}
task1 = task1.add_dataset(dataset1)
task2 = task2.add_dataset(dataset2)
model = GeneralistModel()
Trainer().fit(model=model, tasks=[task1, task2])
\end{pythoncode}
The \concept{GeneralistModel} has various kinds of implementations, which are discussed in \cref{sec:design}.
%For the detailed introduction to training API, please refer to \cref{sec:training}.

Such an instruction interface also enables zero-shot inference with generalist model checkpoints, which allows one to evaluate its generalization ability on unseen tasks:
\begin{pythoncode}
instruction = '[IMAGE:img] what does the image describe? -> [TEXT:cap]'
data = {'img': 'image_1.jpg'}
output = GeneralistModel.from_pretrained('multitask.pt')\
         .inference(instruction, data=data)
\end{pythoncode}

The term \concept{declarative} indicates that a task is created by its formulation rather than the control flow. Researchers can fix the provided model set while only study task-scaling problems, or fix the task set to study the underlying model structure. 
Different from task-specific libraries~\cite{wolf2020transformers,chen2019mmdetection}, \ofasysstyled{} emphasizes on using the same universal model to handle both pretraining and finetuning tasks.

\section{User Interface}
\label{sec:instruction}
% The user interface in \ofasysstyled{} can be divided into the declarative part and the imperative part.
% The declarative part is the instruction interface that is used to define the tasks and the configuration of the whole system.% (see \cref{sec:instruction,sec:training,sec:inference}).
% The imperative part completes the library codes with user-provided implementation and customization (see \cref{sec:extend}).

% \subsubsection{Declarative Task Definition: Instruction}
% \label{sec:instruction}

%\ofasysstyled{} proposes a declarative interface, named \concept{instruction}, to describe a multi-modal task.
%Through instructions, \ofasysstyled{} decouples the user-side task definition from the library-side task implementation.
Now we introduce the details of the instruction user interface, including its formulation and several instruction examples that illustrate the purpose of each consideration.

\subsection{Core Concepts}

\label{sec:grammar}

To allow tasks of diverse forms and multi-modal data of diverse organizations, \ofasysstyled{} supports the following formulation of instructions (expressed in regular expressions):
% The grammar for the instructions (expressed in the manner of regular expressions) is formalized as 
%%%%%%%%
\begin{plaintext}
   instruction:  sentence->sentence
      sentence:  (plain_text|meta_slot)+
     meta_slot:  slot|expanded_pattern
          slot:  \[type(:name)?(,attribute)*\]
     attribute:  (key(=value)?)*
name,key,value:  [_A-Za-z0-9]+
\end{plaintext}
%%%%%%%%

%\paragraph{Instruction}

% As a \concept{instruction} expresses the content of a job, its format naturally should be able to convey the intent of the tasks the job supposed to conduct, as diverse as possible.
% Considering the balance between understanding and generation tasks and models, the instruction follows a sequence-to-sequence normative.
An \textbf{instruction} consists of two \concept{sentences} connected by an arrow ``-\textgreater''.\footnote{The arrow is neither the input nor the output to the system. It is for the system to identify encoder slots from decoder slots.}
% The sentence before the separator specifies the content of the encoder and vice versa.
A sentence contains multiple segments, each of which is either \concept{plain texts} describing the task goals or a \concept{meta\_slot} describing the multi-modal data.
% It is required that each sentence contain at least one data\_spec.
A \concept{meta\_slot} can be either a \concept{slot} or an \concept{expanded\_pattern} consisting of slots. 

%Although each sentence with content must contain at least one slot, it is permitted to use empty sentence to represent an encoder only or a decoder only model.\todo{this is inconsistent with the grammar, sentence can only have plain\_text and sentence cannot be empty}

%\paragraph{Slot}
A \textbf{slot}, identified by square brackets from plain text segments, is the basic processing unit of \ofasysstyled{}.
A slot consists of a \concept{type}, a \concept{name}, and optional \concept{attributes} of \concept{key=value} pairs.
\ofasysstyled{} uses these metadata to configure the modality-aware data processing.
The type specifies the modality of the data.
The name is used to retrieve data from the data source.
The attributes customize data processing.

% \paragraph{Expanded Pattern}
The \textbf{expanded pattern} allows the system to register more syntax on the instruction. There are some typical patterns in \ofasysstyled{} working in progress, \eg, the \concept{interleaved\_pattern} and the \concept{contrastive\_pattern}.

\begin{plaintext}
   expanded_pattern:  interleaved_pattern|contrastive_pattern
interleaved_pattern:  \[slot+\]\*
contrastive_pattern:  \[slot\|slot\]
\end{plaintext}
The interleaved\_pattern is signaled by brackets and a following asterisk and can contain multiple inner slots.
It provides theoretical support to aligned multi-modal data of variable lengths.
% \eg, bounding boxes and their category labels in object detection, and exemplars in in-context learning.
The actual number of inner slots in the computation is determined by the data rather than the instruction. The contrastive\_pattern provides syntax support for contrastive learning.

\subsection{Instruction Examples}

In the following, we demonstrate several representative tasks written by instructions, to help understand how these concepts work in practice.
%and showing how attributes and custom slot helps control the modality processing.
%of what can be achieved using instructions.

\paragraph{Slot Type and Name.}
A slot is only attached with one unique modality, whose type is represented as the slot type. The slot name is used  mainly for field mapping in dataset.
\begin{illus}[Basic Image Captioning]{illus:basic}

\begin{inst}
[IMAGE:img] what does the image describe? -> [TEXT:cap]
\end{inst}

\noindent
The two sentences separated by \concept{-\textgreater} describe the task input and its desired output, respectively. 
%The sentence before is for the encoder and the sentence after, the decoder.
%An instruction normally consists of two sentences separated by \concept{-\textgreater}. The sentence before is for the encoder and the sentence after, the decoder.
In this case, ``[IMAGE:img]'' specifies that there is an image input bound to a data column named \param{img} in the dataset. 
The plain texts in the instruction indicate the task is about captioning an image.
The output of the task is a text sequence, which is the \param{cap} column in the dataset.
\end{illus}

\paragraph{Attributes.} 
Attributes allow  fine-grained control over a certain slot. Users can exploit built-in attributes or implement customized ones.
%An example is shown in \cref{illus:attr}.

%For example, it may be helpful to repeat the source sequence in the target sequence as a prompt and constrain the generation space during training and inference for understanding tasks in the sequence-to-sequence regime. 
% It is commonly seen in open-ended classification, question answering, and detection tasks, 

%%%%%%%%
\begin{illus}[MNLI with Prompt Prefix]{illus:attr}

\begin{inst}
can text1 [TEXT:s1] imply text2 [TEXT:s2]? -> can text1 [TEXT:s1,no_loss] imply text2 [TEXT:s2,no_loss] ? [TEXT:label,closed_set]
\end{inst}

\noindent 
For text classification tasks, \eg, MNLI~\cite{williams18mnli}, we find it helpful to repeat the source as prompt prefix in the output~\cite{wang22ofa}. 
However, the decoder slots default to the cross-entropy loss, which can be disabled using shortcut \param{no\_loss}.
The prefixes are also ensured to be generated by \ofasysstyled{} in inference.
MNLI also has a limited label space, which can be constrained using the attribute \param{closed\_set}. Enumeration of the closed set is specified in the task configuration.
\end{illus}
%%%%%%%%

\paragraph{Variable-Length Slots.}
It is not uncommon for a task to generate several outputs of the same type, \eg, object detection in computer vision.
\ofasysstyled{} can support such type of tasks with variable-length slots using the interleaved\_pattern.
% %%%%%%%%
\begin{illus}[Object Detection with variable-length slots.]{illus:length}
\begin{inst}
[IMAGE:img] detect the objects in the image. -> [[BOX][TEXT]]*
\end{inst}
The output in object detection is normally a bounding box with its label. 
However, an image may contain several objects, whose number is not known in advance.
% \todo{why doesn't it use name to map to data? is it in inference? why is it a asterisk not a plus sign?}
\end{illus}
% %%%%%%%%

\paragraph{Custom Slot.}
One can define their own type of slots to do interesting stuff.
The example in \cref{illus:custom} showcases the versatility of the slot concept. 
% Please be aware that it is only meant as an example and not implemented in \ofasysstyled{}.
For guidance on how to support a new type of slot, please refer to \cref{sec:extend}.
% As a slot is defined as a specific type of data processing, one is allowed to define a slot type even without data, which is instrumental in describing a finetuing job of the prompt-tuning type.
% An example is given in \cref{illus:custom}.
%%%%%%%%
\begin{illus}[Image Captioning with Prompt-Tuning.]{illus:custom}
\begin{inst}
[IMAGE:img] [PROMPT:pt,len=100,prefix-tuning] -> [TEXT:cap]
\end{inst}
% \todo{is it supported to have two prompt slots? and why is the attribute key is prefix-tuning?}
The instruction specifies a job with the customly-defined \concept{PROMPT} slot type.
\ofasysstyled{} registers \concept{len=100} learnable embeddings to each layer using the name \concept{pt}, appends them to the source sequence for each example, and trains them using \concept{prefix-tuning}~\cite{li20prefix}
\end{illus}
%%%%%%%%

The preceding examples demonstrate the generality and versatility of the proposed user interface, which enables expedient task prototyping. 
%\zc{To see more advanced instruction features, \eg, dynamic number of slot composition, position alignment between multiple slots, please refer to Appendix~\ref{appx:tasks}.}
%
Overall, the instruction formulation (together with appropriate implementation) allows the expression of diverse task paradigms.
In theory, it can not only describe tasks of CLM, S2S, and DDPM paradigms shown in \cref{tab:comparison}, but also support paradigms such as soft prompt tuning in NLP~\cite{li20prefix} through custom slots, CLIP-style contrastive learning~\cite{radford21clip} through custom criteria, and Flamingo-style in-context learning~\cite{alayrac2022flamingo} with custom instruction parsing. %, the implementation of which is work in progress.

% \reviewcheck{The following is supposed to show how to do things with OFAsys, such as training, validation, ..., you can treat it as a readme or short tutorial, or run training in 5 minutes. the introduction of related components is naturally incorporated.}

%%%%%%%%
\begin{figure}[t]
    \centering
    \includegraphics[width=0.8\linewidth]{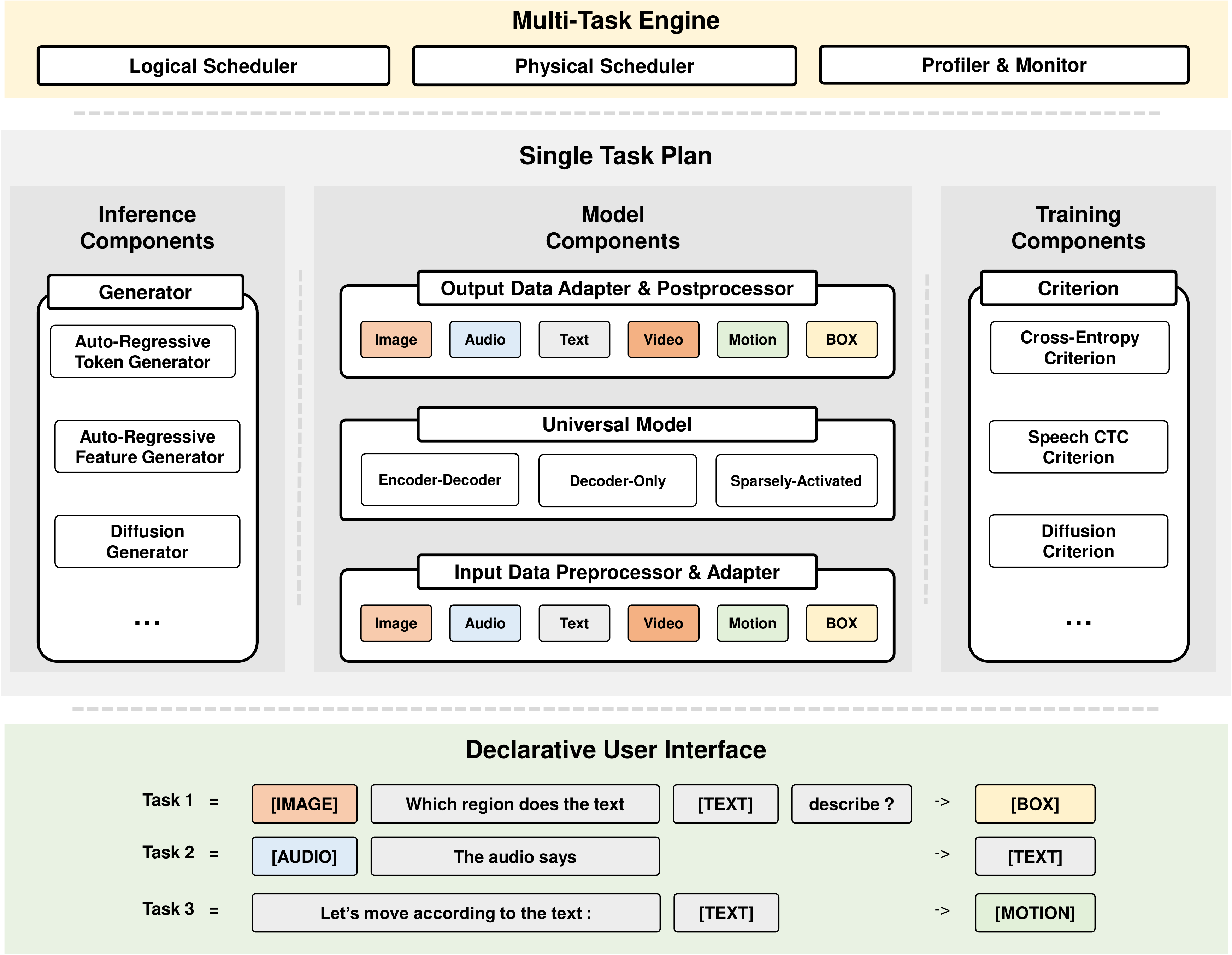}
    \caption{Overview of main components and their organization in \ofasysstyled. %\ofasysstyled{} provides 
    The declarative user interface is realized with a hierarchy of reusable components. 
    Based on instructions, \ofasysstyled{} automatically generates plans for single task execution in both training and inference. Moreover, the multi-task engine provides support to efficient multi-task learning. %, including how each task should be performed, trained, evaluated, and inferred. %\ofasysstyled{} offers a wide range of built-ins, in terms of data adapters, model plugins, training objectives, and inference generators, to deal with multi-modal multi-task learning. 
    }
    \label{fig:overview}
\end{figure}
%%%%%%%%

\section{System Design and Implementation}
\label{sec:design}

% In this section, we demonstrate the system design  as well as the implementation on top of the multi-modal instruction interface.
In this section, we start from the motivation of the system design with its relation to the instruction interface, and further introduce the details of the system implementation to reveal the considerations in realizing the design.

% We start from the motivation of this study, and further introduce the details of system design to unveil the secrets behind the simple instruction interface. 

\subsection{Scaling Challenges in Training Multi-Modal Multi-Task Models}
\label{sec:motivation}
% 个人认为motivation得上来就讲，肯定多少和intro有些重，但主要是让读者看这个design能从认可你的立场开始往下看。而且很多人可能看design的可能就不怎么看讲故事的intro

% An obvious question is why we need the disentangled components.
% Let's imagine a common scenario in machine learning field.  researchers want to explore the ability of an existing model on a new task. 
% The success of pretrained models has promoted its extensive application to novel tasks in diverse fields.
% However, a prevalent difficulty in introducing such methods to a new task is that complex procedures are involved, \eg, creating a dataset ready for training, designing the model structure for the adaptation to new data or new modality, elaborating criteria for effective training.

% A common painful experience in today's deep learning practice is suffering from the complex procedures of building and training a model on a single task. 
% What researchers need to do includes but is not limited to: creating a new dataset and getting it ready for training, designing the model structure for the adaptation to new data or new modality, elaborating criterions for effective training, etc. 
% Recent frameworks such as Transformers, Fairseq, ModelScope, etc., greatly reduce the complexity for users, and have essentially facilitated single-task learning. 

%\zc{The above paragraph shrink to one sentence. and then starts with sth like, \ofasysstyled{} provides such convenience to build a multimodal task workflow from a slot-centric or modality-centric view, which is a more standarlized procedure.} 

Frameworks such as fairseq~\cite{ott0219fairseq} and transformers~\cite{wolf2020transformers}, have \textit{de facto} standardized and streamlined the procedures of a specialized group of deep learning methods, reducing the development cost of handling task-specific training and inference.
However, these frameworks are not sufficient with the surge of multi-modal and multi-task learning~\cite{flan,sanh22t0, natural_instruction_v2,wang22ofa,gato, flamingo,lu2022unifiedio}, which faces profound challenges in terms of the heterogeneity of multi-modal data, the diversity of task formulation, and the complexity of scalable schedule of computation.
Researchers have to, on their own, (a) implement specific data processing procedures for every task, (b) adapt the model structure and computation to each task with different feature extractors and losses, (c) manually determine the task precedence during optimization for  the model performance, and (d) manage sample batching, op-level replacement of highly diverse task workloads in distributed environment, for training efficiency. 

% However, with the surge of multi-task learning~\cite{flan, sanh22t0, natural_instruction_v2} and unified multi-modal pretraining~\cite{wang22ofa, gato, flamingo, lu2022unifiedio}, 
% there is a necessity for an AI system to manage tasks and modalities more efficiently.
% It is insufficient for the existing frameworks to reach the objective, due to the heterogeneity of modalities and the scale of tasks. 
% Thus, researchers have to spend much efforts on the aforementioned procedures task specifically, and also have to manually schedule the tasks concerning multiple modalities. 
% This severely impedes the development of the generalist model, which calls for liberation from the tedious work. 

To facilitate multi-modal multi-task learning towards generalist models, we develop the One-For-All System (OFASys) to seamlessly integrate multiple modalities and multiple tasks into a single framework with universal models, as well as automatic task schedulers to manage multi-task execution. 

\subsection{System Overview}
% 是不是直接一个data flow开场，用计算过程来表达整个系统的实现，更加容易理解呢？
We illustrate the overview of the system with an example of the data flow of a single task to show how heterogeneous inputs are processed by \ofasysstyled{}, and further illustrate the design on managing multiple tasks \wrt to single task plans. 

\textbf{Single Task Plan.}\hspace{.5em} 
\ofasysstyled{} accesses the task definition and task data through instructions, as introduced in \cref{sec:instruction}.
% The access of data to the system is the \textit{Instruction}, as demonstrated in Sec.~\ref{sec:instruction}. 
To realize the instruction as viable computation, \ofasysstyled{} parses the instructions into task plans\footnote{We borrow the term \textit{plan} from the \textit{logical/physical plan} in database literature~\cite{databasebook}.}. 
In each plan, there is a model hierarchy, consisting of modality-specific preprocessors/postprocessors and adapters, as well as a modality-agnostic computation model. 
For an example, the system first retrieves data specified by the slots and dispatches them to corresponding preprocessors to covert them into common ML input types, \eg, tokens for texts and fbank features for audio.
% Through slots in the instruction, the system parses the data and dispatches them to the corresponding preprocessors for data transformation, \eg, converting to tensors.  
Then, the preprocessed data are dispatched to the corresponding adapters for feature extraction, and the output features are joined to form sequences of representations for the universal model.
These steps ensure a unified data format for the universal model but with reusable and composable data processing pipelines.
% After the transformation, data are dispatched to the corresponding adapters for feature extraction also based on their slot information. 
The universal model is namely a general module for fusing multimodal inputs and generating outputs.%, and it plays as the core of the whole system. 
As the inputs and the outputs are consistently being representation sequences, the implementation of the universal model is highly versatile, regardless of the modality intricacies.
The outputs of the universal model is finally postprocessed by the adapters and postprocessors, in order to generate content consistent with the input formats. %, \eg, discrete tokens, feature vectors, etc. 
Stage-wise components, including criteria and generators, provide support in training and inference, which have a variety of out-of-the-box implementations. 
In this way, different multi-modal data can go through the system with consistent inner interfaces to improve development efficiency.

% The above components construct a single-task plan. 
\textbf{Multi-Task Plan.}\hspace{.5em}
In multitask learning, there are multiple such plans parsed from the instructions.
There are two problems to be dealt with: (a) how a single model is used with multiple plans and (b) how  multiple tasks are optimized and executed together.
For the first problem, \ofasysstyled{} shares the trainable parameters of the adapters and the universal model by default, such that each parameter can be optimized on as many examples as possible.  
To be specific, the modality-specific components are shared across the same modality and the universal model is shared across all tasks.
For the second problem, %the system is further required to manage the tasks for efficient task collaboration. 
%In \ofasysstyled{}, 
the task scheduler is in charge of the management of task precedence and the optimization of the execution details of multi-task workloads with two levels of abstraction.
\ofasysstyled{} automatically combines all the plans to form a multi-task plan with a \textit{logical} scheduler and then the workflow is arranged on physical devices with a \textit{physical} scheduler.
More details are given in \cref{sec:scheduler}.

\subsection{Slot-wise Multi-Modal Data Processing for Composable Task Definition}

%%%%%%%%
\begin{table}[t]
    \centering
    \scriptsize
    \setlength{\tabcolsep}{3pt}
    \begin{tabular}{@{}l l c c l l@{}}
        \toprule
        Slot Type                       & Adapter                       & Encoder & Decoder & Preprocessor                                       & Postprocessor                                    \\\midrule
        \multirow{2}{*}{\texttt{TEXT}}  & Embedding Lookup (Text)       & \cmark  & \cmark  & text$\rightarrow$token                             & text$\leftarrow$token                            \\
                                        & Embedding Lookup (Phone)      & \cmark  & \xmark  & text$\rightarrow$phone                             & N/A                                              \\\hline
        \multirow{2}{*}{\texttt{IMAGE}} & Vision CNN/ViT                & \cmark  & \xmark  & image$\rightarrow$image3d                          & N/A                                              \\
                                        & Embedding Lookup (Image Code) & \cmark  & \cmark  & image$\rightarrow$code                             & image$\leftarrow$code                            \\\hline
        \texttt{VIDEO}                  & Vision CNN                    & \cmark  & \xmark  & image$\rightarrow$video4d                          & N/A                                              \\\hline
        \texttt{AUDIO}                  & Acoustic CNN                  & \cmark  & \cmark  & wave$\rightarrow$fbank                             & wave$\leftarrow$fbank                            \\\hline
        \texttt{MOTION}                 & Linear Projection             & \cmark  & \cmark  & BVH$\rightarrow$motion6d                           & BVH$\leftarrow$motion6d, GIF$\leftarrow$motion6d \\\hline
        \texttt{BOX}                    & Embedding Lookup (Box Code)   & \cmark  & \cmark  & bounding box$\rightarrow$code                      & bounding box$\leftarrow$code                     \\ \hline
        \texttt{STRUCT}                 & Embedding Lookup (Text)       & \cmark  & \cmark  & schema$\rightarrow$token, sudoku$\rightarrow$token & sudoku$\leftarrow$token                          \\
        \bottomrule
    \end{tabular}
    \caption{Modality support in terms of slot types in \ofasysstyled.}
    \label{tab:support-matrix}
\end{table}

As mentioned in \cref{sec:motivation}, with existing frameworks, one has to organize the data for preprocessing differently \wrt each task, which is laborious and time-consuming. %heavy work in practice. 
To address this issue, especially for multi-modal data, \ofasysstyled{} introduces slot-wise multi-modal processing realized by modality-specific preprocessors, postprocessors, and adapters.
They conduct data transformation and feature extraction on a modality basis, which can be easily composed to define new multi-modal tasks.
% In this context, we include both preprocessor and adapter, which serve modality-specific data transformation and feature extraction, to this procedure.
% Differently, to reduce these efforts, \ofasysstyled{} uses slots as the basic input units for data processing. 
% In this way, users can specify modality and data with slots in the instruction without worrying about the following procedures. 
To be specific, \ofasysstyled{} includes a dispatcher that automatically sends the data in each slot to the corresponding preprocessor/postprocessor and adapter, according to the slot type, \ie, its modality.
The relation between slot types and slot-wise components is shown in \cref{tab:support-matrix}.

Take Task 1 in \cref{fig:overview} as an example. 
The instruction is parsed as 5 slots: ``[IMAGE]'', ``Which region does the text'', ``[TEXT]'', ``describes ?'', and ``[BOX]''. 
The first slot is processed by the image-specific preprocessors and adapters. 
The three ones after essentially represent texts, and they are processed by the text-specific ones. 
The last one is processed by the ones for bounding boxes. 
By default, \ofasysstyled{} dispatches data to the processors of the corresponding modalities. 
This liberates users from the manual work on task-level modality-specific data processing. 

% The middle three slots will be handled by the preprocessor and adapter for the TEXT modality.
% The "[IMAGE]" slot and the "[BOX]" slot are also handled by corresponding modality-specific components.
% More specifically, we divide the process of converting data from the original form into a form readable by the \concept{Compute Engine} into two steps, and use two modules, \concept{preprocessor} and \concept{adapter}, to process them respectively.
% These preprocessors and adapters can be reused between multiple tasks.
% All the user needs to do is specify the preprocessor and adapter used by each slot in the instruction.

\textbf{Preprocessor and Postprocessor} provide the ability to convert data from raw format to common machine learning data types and vice versa.
Preprocessors take raw data as input, convert them into the proper form, and then collate multiple data example in a mini-batch into batched data, preparing for subsequent batch processing.
Postprocessors convert model outputs to the original input format.  
In most cases, a postprocessor perform an inverse process of the corresponding preprocessor.
For multiple slots of different modality in an instruction, \ofasysstyled{} includes a \concept{dispatcher} to better manage the process.
The dispatcher first applies the assigned preprocessor to the input data in each slot, and then the slots are processed in groups to improve efficiency.
The slots from different examples are finally collated for batch processing in terms of their order in the instruction, \eg, the data in the first image slot in the instruction from given examples are collated.

% general preprocessor 这个名字可能要改叫planner之类
% To better organize the preprocessors for different slots, we build a preprocessor-planner to dispatch slots and collate tensor format data.
% When called, the preprocessor-planner feeds each slot to its assigned preprocessor and call the single-slot Map method in order. 
% The system then concatenates the processed data by modality with the Group-Map method to improve the efficiency in the subsequent batch processing. 
% Slots from different samples are finally collated according to their positions for batching processing.
% Postprocessors serve for converting model outputs to the task-wise output formats.  
% In most cases, postprocessors perform an inverse process of the preprocessors.
% In practice, we usually treat postprocessors as the decoding method in the preprocessor class to make sure they are implemented in pairs.

% Similar to how sensory stimulation is translated into organized experience in the human brain, adapters convert data of different modalities between its tensor format represented by computer programs and embedding sequences accepted by the universal model.
% \eat{Recent advances in self-supervised learning make IO adapters considerable strong for the downstream tasks, showing that unimodal universal representation can well model the joint probability distribution and conditional probability distribution, \eg, MAE model~\cite{mae} can even recover the image when only \qty{15}{\percent} pixels are given.} 
\textbf{Adapter} plays as the role of modality-specific feature extraction or representation learning. 
Adapters build a consistent input/output interface that unifies the difference among modalities for the universal model .
Each \concept{input adapter} takes the preprocessed data in the slot as input and outputs embedding/representation sequences.
Input adapters can also produce auxiliary data needed by the model, \eg, positional embeddings in terms of self-attention biases~\cite{raffel20t5}.
If necessary, \concept{output adapters} are implemented as an inverse process of the input adapters. 
Similar to preprocessors, \ofasysstyled{} includes a dispatcher to manage the process.
The dispatcher applies the adapter to the batched preprocessed input and finally concatenates the representation sequences from slots to form the input of the universal model.
As adapters contain trainable parameters, to avoid unnecessary memory costs in training, the adapters not used in the instructions are not initialized.
%for further modality-specific processing of the outputs.
% Each \concept{input adapter} takes data of a certain modality in the format of tensor from the slot as input, and then output embedding sequences in the same format.
% By default, \concept{input adapter} not only returns the embedding sequences, but also some additional tensors, \eg, positional embeddings in terms of self-attention biases~\cite{raffel20t5}. 
% Similar to the preprocessor-planner, we build an adapter-planner to dispatch the processed data and concatenate their outputs.
% To avoid unnecessary memory costs, the adapter-planner only activates the adapters declared in the instruction.
% In the forwarding process, the adapter-planner applies each slot to the \textit{input adapter} in order of the modality and concatenate the embedding sequences  as the return. 
% The outputs of the adapter are then processed by the compute engine. 
% If necessary, the output adapters are implemented for further modality-specific processing of the outputs of the engine. 
% If necessary, the adapter-planner will split the whole model output into the model output of slots, and then use the output adapters to convert the model output of each slot back to raw data format with the help of postprocessors. 
% Adapters are also implemented as paired methods in a class.

In existing application frameworks, the adapters in \ofasysstyled{} are often considered as part of the static overall model. 
However, adapters are by nature modality-specific and such practice limits the flexibility of multi-task learning on new modalities and new tasks. 
Therefore, \ofasysstyled{} disentangles modality-specific model computation from the universal backbone model computation.
Yet a new problem arises that preprocessors/postprocessors and adapters are both modality-specific and one may implement data processing in either component. 
The difference in design is that adapters work on batched data and contain trainable parameters, which are saved in checkpoints.

\subsection{Modality-Agnostic Computation with Unified Representations}

As the modality-specific data processing resolves the difference among modalities, the core model computation, namely the universal modal, can focus on the architecture design.
\ofasysstyled{} can accommodate various universal model structures, including transformer-based sequence-to-sequence models~\cite{vaswani17attention} like T5~\cite{raffel20t5}, U-Net~\cite{unet} for diffusion~\cite{ho-ddpm}, decoder-only models like GPT~\cite{brown20gpt3, chowdhery22palm}, \etc.
The only requirement for the universal model is that it shall take an embedding sequence as input and outputs another embedding sequence. %, which handles all tasks in a unified and shared way.

% \paragraph{Universal Compute Engine}

% Though there are also works ~\cite{albef,coca,flamingo}
% showing that modal-dependent architectures, i.e., bi-modal cross-attention, can achieve competitive performance,  such designs can not be naturally scaled to more modalities ($>$3). 
% \ofasysstyled{} fuses different modalities through a universal model, which takes a concatenated embedding sequence as input and outputs another.
% The universal model is designed as a black box to optimize all tasks in a uniform and shared way.
% This abstraction disentangles the complexity of multi-modal data processing from the model implementation and thus enables easier modification of the model structure. 
% Relying on the disentangled design, \ofasysstyled{} potentially supports various model structures which are not limited to transformer-based models, such as the U-Net \cite{unet} for diffusion~\cite{ho-ddpm}, Decoder-only models like GPT~\cite{brown20gpt3, chowdhery22palm}, \etc.

We currently provide a transformer-based encoder-decoder model as the default implementation of the universal model. 
Recent progress in different fields have witnessed the potential of the Transformer architecture becoming the universal framework, and sequence-to-sequence learning might be a paradigm towards generalist models~\cite{wang22ofa, gato, chowdhery22palm, flan, flamingo}. 
Both the encoder and the decoder consist of Transformer blocks, each including self-attention, cross-attention, and point-wise feed-forward network (FFN). 
Additionally, inspired by \cite{wang22beitv3, wang20vlmo}, we implement a sparsely-activated Mixture-of-Experts (MoE) model as an alternative. 
\subsection{Stage-wise Components}

% 这段有啥用。。 renxc: 这是的意图应该是说明slot的结果怎么跟model结合到一起的，包括那个中间的universal model和用到这些结果的criterion （preprocessor, output adapter）和generator (input/output adapter)。但没有现成的内容，我就把原来相关的东西粘过来了。
To build a complete computation task in training or inference, \ofasysstyled{} provides several stage-wise components, including criteria and generators. %, to build a complete task.
For each task, commonly-used criterions and generators provided by the system can be specified accordingly in the instruction.
% In each task, commonly-used criterions and generators provided by the system can be freely specified in the instruction by the user.
In the stage of training, the criterion computes the loss using the output of the above components and can be automatically deducted from the instruction.
% The criterion computes the loss using the constructed model and the data.
In the stage of inference, the generator is interpreted from the decoder slot in the instruction, which generates the final output with the help of output adapters and postprocessors.
% The generator produces the output using the constructed model and the input.
% The evaluator computes the metric values using the generated output.

% \todo{bypass the description of functionality, list the supported criterions, it can change the model computation and includes losses for various training objectives.}
% Similar to most deep learning methods, the unified model in \ofasysstyled{} is optimized via stochastic gradient descent using supervised data.
% To obtain gradient of the parameters, an optimization objective is need, which is often implemented as criterions or loss functions in deep learning frameworks.  
% \zc{We waste more than 100 words to say what a criterion or loss function is.}
%\ofasysstyled{} learns models by regular back propagation, and we need to specify loss criterions for a given task.

% We have two different criterion views, a slot-centric view that is slot-specific and a global view that is visible to all slots. 
% By such fine-grained granularity control over different slots, we can simultaneously support auto-regressive models, diffusion models, etc., for different slots.
%A \concept{criterion} in \ofasysstyled{} 
\textbf{Criterion}
forwards data through the model and uses the output of pre-/post-processors and adapters to calculate the loss for training. 
Several commonly used criteria are provided in \ofasysstyled{}. %, and both types of slots may contain optimization objectives. 
For example, CTC loss~\cite{ctc} that is commonly considered in the ASR task for speech models~\cite{speecht5} can be declared on encoder slots.
For outputs that are in the form of discrete tokens, \eg, language tokens and image tokens~\cite{vqvae,esser21taming}, by default, decoder slots
perform softmax cross entropy within each associated vocabulary, similar to a language model. 
% Despite the common use of the language model style loss in \ofasysstyled{}, %\ofasysstyled{} also makes it possible to enable different loss functions for different slots. 
% %For example, 
% it theoretically supports both MSE for image pixels and inner batch contrastive learning for implicit vocabulary~\cite{wav2vec2.0} on E-slot.
% However, as D-slots, in addition, affect the generator in inference and contain more control flows of the implementations, implementing other criterions on D-slots is not trivial.
Besides token-level teacher forcing with the cross entropy criterion, the system provides other flexible learning paradigms, such as sequence-level loss,  which supports reinforcement learning for action-reward tasks, and denoising diffusion probabilistic modeling (DDPM)~\cite{ho-ddpm}, one of the most commonly-used diffusion methods.

% \subsubsection{Inference Components}

% In inference, the generator and the evaluator are interpreted from the decoder slot in the instruction.
% The generator produces the output using the constructed model and the input.
% The evaluator computes the metric values using the generated output.

% As \ofasysstyled{} follows the encoder-decoder architecture, the output is mainly produced by the decoder. 
% However, as the decoder accepts both E-slots and D-slots of various slot types, different generators are needed to complement the differences in generation paradigm.
%\ofasysstyled{} utilizes generators to produce the final output.
\textbf{Generator}
is used to produce the final output using the model and provided data in inference.
%
%There are three generators in \ofasysstyled{}.
%The generation category task is the main purpose of \ofasysstyled{}. So a unified generator API is designed to enable the model to derive modality-specific outputs. 
Generators in \ofasysstyled{} are separated into two categories according to the generation paradigm, \ie, the \concept{auto-regressive generator} and the
\concept{diffusion generator}~\cite{ho-ddpm}. 
The auto-regressive generator takes an auto-regressive approach to generation, which can be further divided into the discrete token generator and the continuous feature generator.
% The former produces discrete codes,  while the latter generates continuous features.
The diffusion generator performs generation in a non-auto-regressive manner, where the model iteratively denoises the input to derive the output.
The provided implementation covers all the supported modalities in generation.

\subsection{Flexible and Efficient Multi-Task Training with Schedulers}

\label{sec:scheduler}

% The multi-task engine can be conceptually divided into (a) the schedulers, which schedule the tasks in terms of logical optimization steps and physical devices, and (b) the profiler \& monitor, which records the training progress.

% \subsubsection{Schedulers}
% %%%%%%%%%%%%%%%%%%%%%%%%%%
% \begin{figure}
%     \centering
%     \includegraphics[width=.8\linewidth]{fig-schedule-crop.pdf}
%     \caption{A conceptual illustration of tasks in terms of logical plans and physical plans. The directed diagram represents the computation graph of the model. Each node is an individual computation module including preprocessors, postprocessors, adapters, encoders, and decoders. The nodes of the same color support the same type of data. The nodes with dashed edge are inactive for that task.}
%     \label{fig:schedule}
% \end{figure}
% %%%%%%%%%%%%%%%%%%%%%%%%%%
A critical challenge in multi-task learning is the scheduling of the tasks. 
In conventional practice, users need to first implement the aforementioned procedures of data and model processing task specifically, and then manually organize the training of multiple tasks with heuristic rules. 
% For example, \cite{wang22ofa} manually set up a ratio for different tasks in the training with task collaboration, which is fully based on human experience. 
For example, in implementation,  \cite{wang22ofa} mixes the logic of all tasks in the code. 
If users want to add more tasks, they must be careful with the logic coupled with the previous tasks.
% The requirement in this scenario is an efficient task scheduler for fast experimentation. 
The requirement in this scenario is the decoupling of task definitions and an efficient scheduler to train them together.

%To be more specific,
In \ofasysstyled{}, a task scheduler is responsible for training a model with multiple tasks.
%in one optimization step.
The execution of multi-task training is conceptually divided into two levels of abstractions in \ofasysstyled{}:
the \textit{logical scheduler} manages the strategy to compute the overall loss among tasks for each optimization step, while the \textit{physical scheduler} determines how to partition and place the whole computation graph to physical devices with limited distributed capacity.
%They are realized using two schedulers: the logical  scheduler and the physical scheduler, respectively. 
% \cref{fig:schedule} illustrates the goal of the two schedulers.
% The abstraction of logical \vs physical schedulers allows one to compose their own schedules more easily.

For the logical scheduler, users can implement their own strategies, such as those in multi-task learning~\cite{zhang2021mtlsurvey}  or continual learning~\cite{de2021continual} literature, to decide either the task optimization order or the task importance. The default implementation provided by \ofasysstyled{} is a  weighted average of all task losses, where users can adjust the task weight manually.

For the physical scheduler, efficient multi-modal multi-task training in distributed environments is highly challenging~\cite{barham2022pathways}. 
To illustrate, each task may activate a sub-part of the whole model, especially for the I/O adapters; the batch size and the sequence length of each task can also be different, resulting in uneven workloads between tasks. 
%The gradient accumulation--based scheduler is the default in 
\ofasysstyled{} starts with a gradient accumulation-based scheduler. 
In one optimization step, each task performs forward and backward processes separately and accumulates gradients in the local device. 
Then, all devices reduce the gradients and update them for a step of optimization. 
Its advantage is that with the same configuration of each task, such as batch size and sequence length, the peak memory occupation of GPU devices can be the same as in single-task training.
This implementation also supports each task using a different batch size.

\section{Application Example: The OFA+ Models}
\label{sec:application}

%which conveniently used to scale tasks and modalities, 
%We verify the performance of 23 single tasks written by the \ofasysstyled{} instruction interface.  
We train the generalist models, referred to as OFA+~(Generalist) and OFA+~(Generalist+MoE), which can handle text, image, speech, video and motion data all-in-one for the first time, using \ofasysstyled{}. 
With the help of such a system, we conduct easy and fast investigation on multi-modal task-scaling as well as model structure compositions with reliable performance.

%The OFA+ models, which spans over 7 modalities and be trained on 17 tasks, can induce a better performance than UnifiedIO. %Do not mention Unified-IO here, it will make our statement weak.
%Based on \ofasysstyled{}, we can conveniently explore the performance difference brought by only changing the model structure, such as adding MoE, with the same formulation and no need for task-specific customizations or parameters.
%For more applications, including results on single tasks, please refer to \cref{appx:results}.

\subsection{Settings}

Apart from models finetuned on each of the task, the OFA+ (Specialist) models, we train two versions of generalist models as the validation of the system design and implementation which we call OFA+ (Generalist) and OFA+ (Generalist MoE).

The first model, OFA+ (Generalist), is of similar structure to OFA-base~\cite{wang22ofa} and is initialized from the pretrained checkpoints of OFA-base.
It has \numm270m parameters in total, of which \numm90m are modality-specific parameters.
The second model, OFA+ (Generalist MoE), is also based on OFA-base.
Especially, the model structure is augmented with a sparsely-activated implementation of the universal model.
An existing similar implementation is VLMO~\cite{wang20vlmo}, which distributes FFN in transformer layers based on image and text at the bottom layer and visual language at the top layer.
Different from VLMO, we distribute FFN based on different modalities of the slot, such as text, image, speech, video, and motion, at each layer in the encoder.
% Unlike VLMO \cite{wang20vlmo}, which distributes FFN based on image and text at the bottom layer and visual language at the top layer, we distribute FFN based on different modalities of the slot, such as text, image, speech, video, and motion, at each layer in the encoder.
%With the plugin, 
There are \numm455m parameters in total, of which \numm275m are modality-specific.

\begin{figure}[t]
\begin{minipage}{.48\linewidth}
%%%%%%%%
% \begin{figure}[t]
    \centering
    % \hfil
    % \begin{subfigure}{0.48\linewidth}
        \includegraphics[height=13\baselineskip]{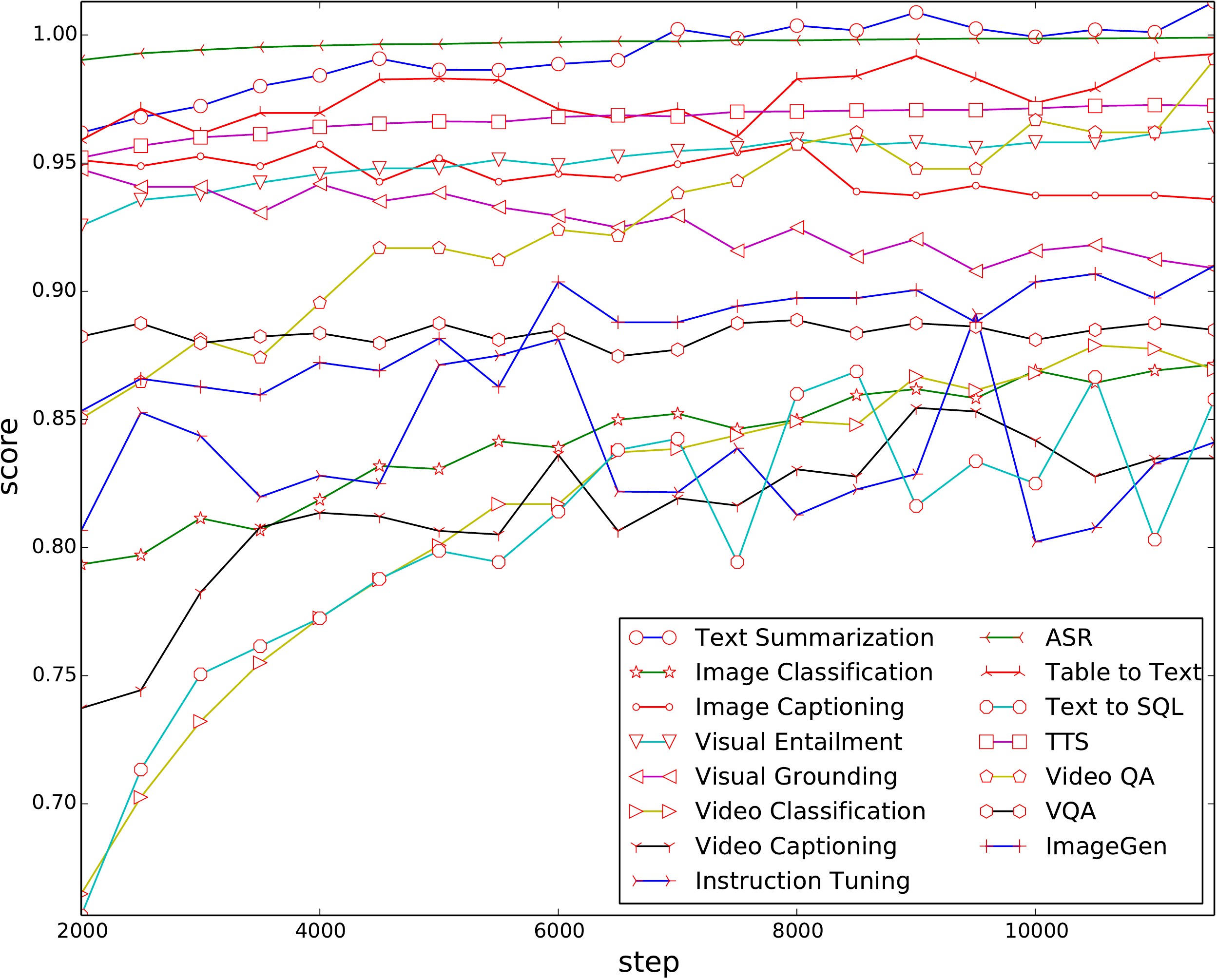} 
    % \caption{OFA+ (Generalist)}
    
    % \end{subfigure}
    % \hfil
    % \begin{subfigure}{0.48\linewidth}
    %     \includegraphics[height=14\baselineskip]{fig-moe-crop.pdf} 
    % \caption{OFA+ (Generalist) \vs OFA+ (Generalist MoE)}
    % \label{fig:moe}
    % \end{subfigure}
    % \hfil

    \caption{Learning curves in terms of metrics for OFA+ (Generalist). %and OFA+ (Generalist MoE). 
    The y-axis represents the percentage of the performance \wrt the corresponding specialist. %In order to keep consistent with other tasks, the scores of TTS and ASR are linearly transformed.  
    The results of TTS and ASR are linearly transformed so that for all results, higher is better.
    %The higher, the better.
    As we can see, different tasks have divergent learning speed and reach maximum performance at different steps. 
    %In addition, the MoE version converges faster and better in modalities such as audio and video.
    }
    \label{fig:curves-all}
    \label{fig:learning-curves}
% \end{figure}
%%%%%%%%
\end{minipage}
\hfill
\begin{minipage}{.48\linewidth}
%%%%%%%%%%%%%%%%%%%%%%%%%%%%%%%%%%%%%%%
% \begin{figure}[t]
    % \centering

    \hfil
    \begin{subfigure}{0.46\linewidth}
    \centering
    \includegraphics[height=6\baselineskip]{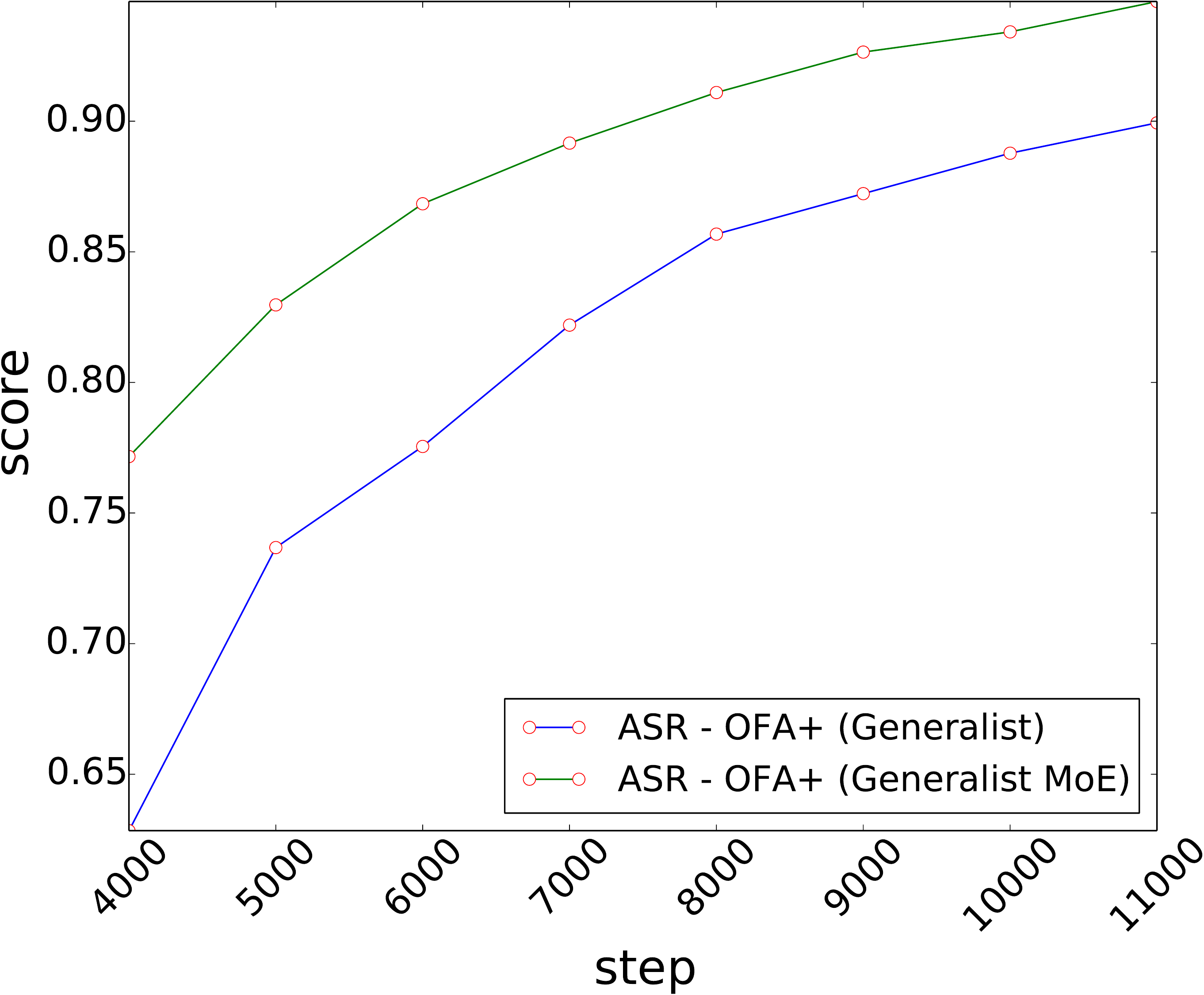}
    \caption{ASR}
    \end{subfigure}
    \hfil
    \begin{subfigure}{0.48\linewidth}
    \centering
    \includegraphics[height=6\baselineskip]{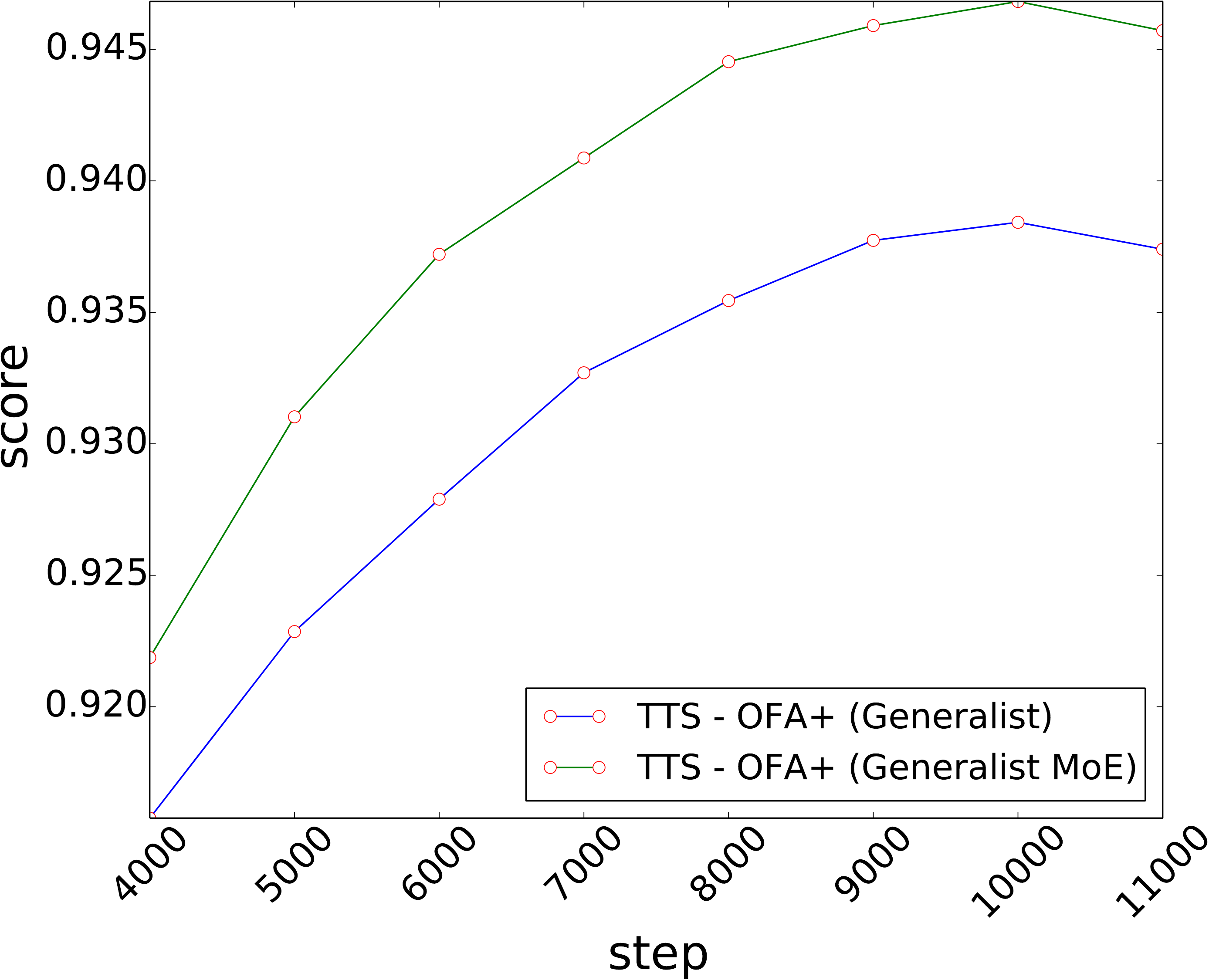}
    \caption{TTS}
    \end{subfigure}
    \hfil

    \hfil
    \begin{subfigure}{0.48\linewidth}
    \centering
    \includegraphics[height=6\baselineskip]{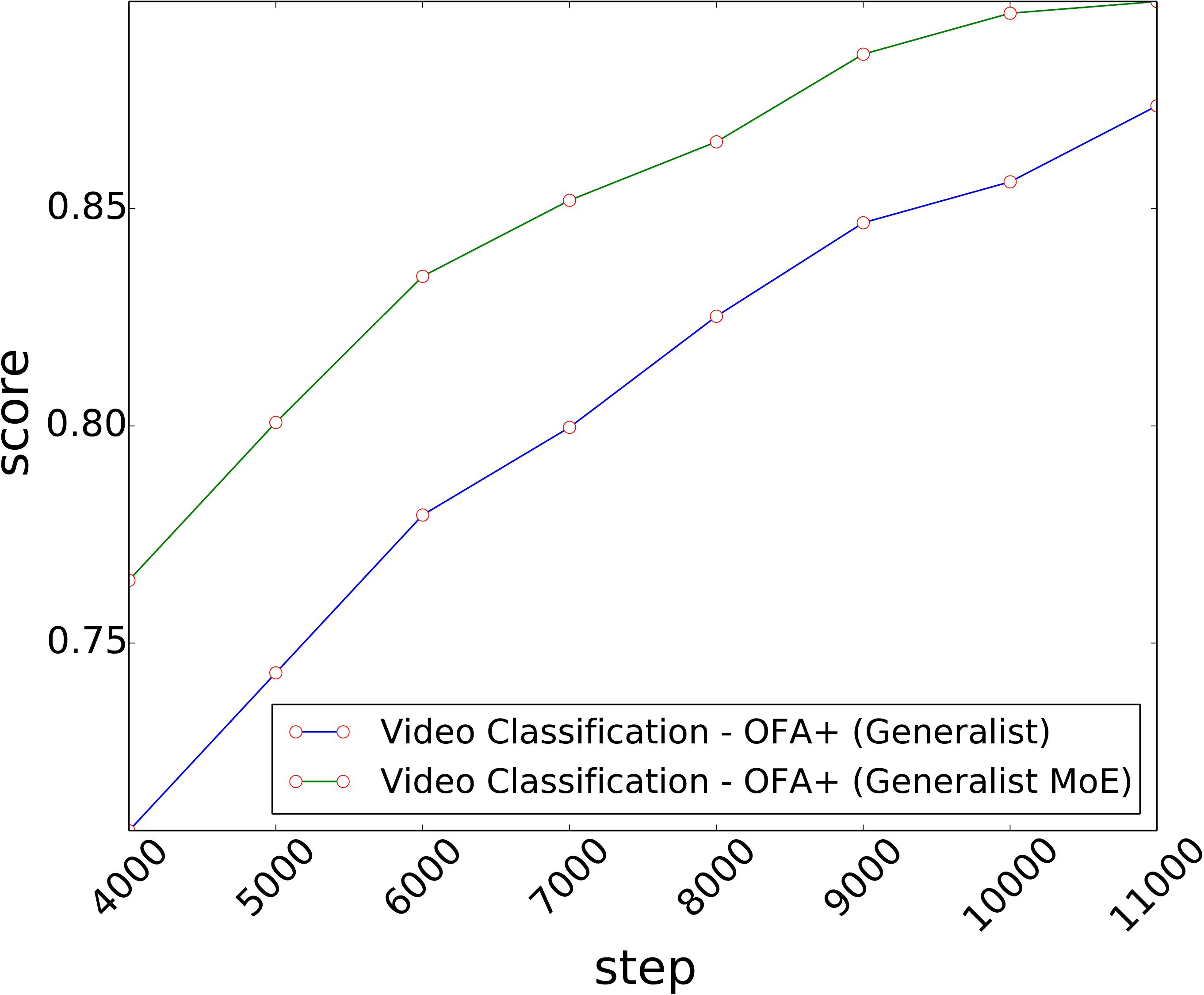}
    \caption{Video Classification}
    \end{subfigure}
    \hfil
    \begin{subfigure}{0.48\linewidth}
    \centering
    \includegraphics[height=6\baselineskip]{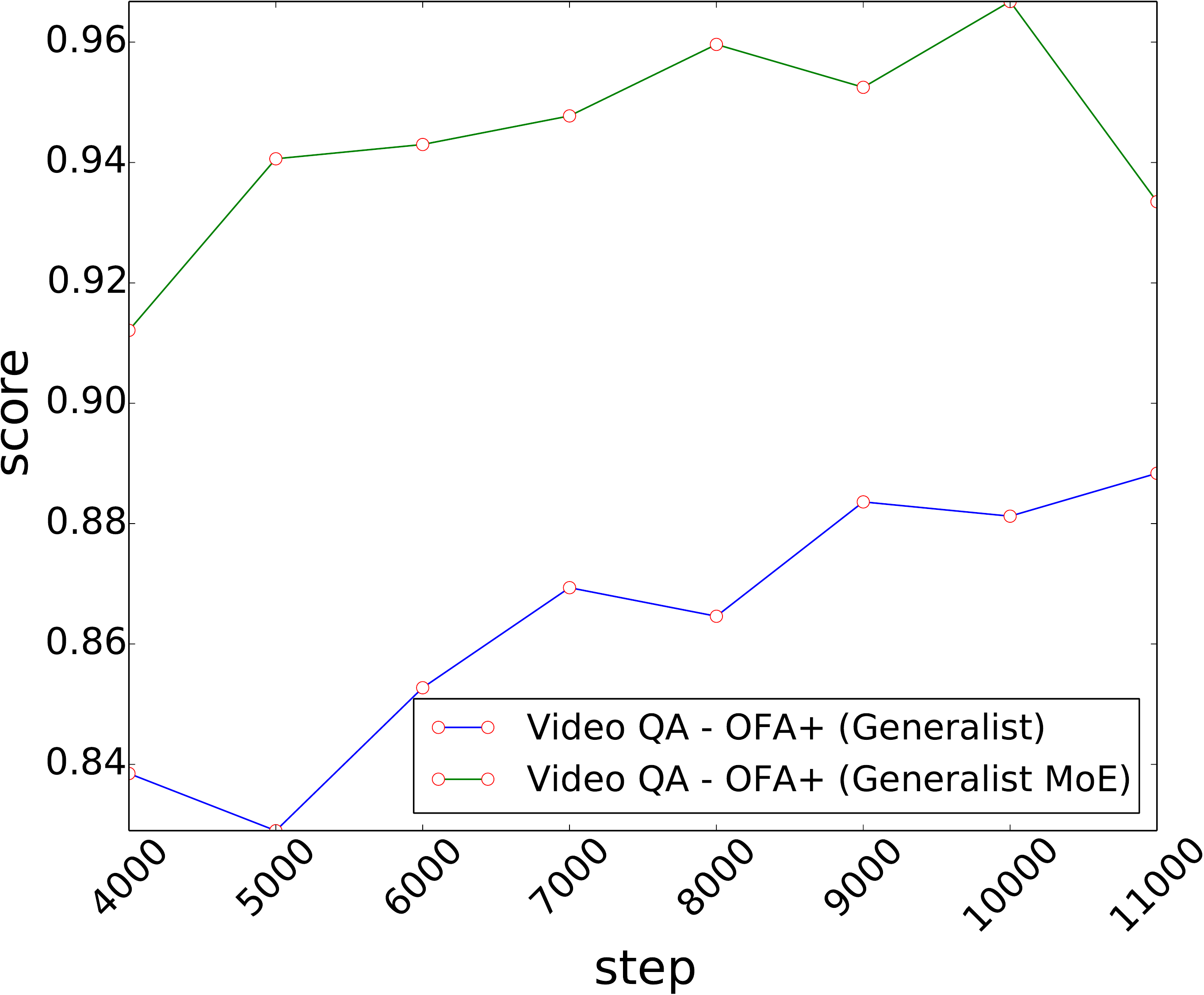}
    \caption{Video QA}
    \end{subfigure}
    \hfil
    
    % \nextfloat
    \caption{Learning curves in terms of metrics for OFA+ (Generalist) and OFA+ (Generalist MoE) on representative tasks.
    Arranged the same with \cref{fig:curves-all}.
    % The y-axis represents the percentage of the performance \wrt the corresponding specialist.
    % The higher, the better.
    %As we can see, different tasks have divergent learning speed and reach maximum performance at different steps. 
    It can be seen that the MoE version converges faster and better in modalities such as audio and video.}
    \label{fig:moe}
% \end{figure}
%%%%%%%%%%%%%%%%%%%%%%%%%%%%%%%%%%%%%%%
\end{minipage}
\end{figure}

% %%%%%%%%
% \begin{figure}[t]
%     \centering
%     \includegraphics[height=14\baselineskip]{fig-multi-task-curves-crop.pdf} 
%     \caption{Learning curves in terms of metrics for OFA+ (Generalist). We find that different tasks have divergent learning speed and reach maximum performance at different steps. The y-axis represents the percentage of the performance \wrt corresponding specialist.}
%     \label{fig:learning-curves}
% \end{figure}
% %%%%%%%%

% %%%%%%%%
% \begin{figure}[t]
%     \centering
%     \includegraphics[height=14\baselineskip]{fig-moe-crop.pdf} 
%     \caption{Learning curves in terms of metrics for OFA+ (Generalist) and OFA+ (Generalist MoE). We find that MoE converges faster and better than Generalist in different modalities such as audio and video. The y-axis represents the percentage of the performance \wrt corresponding specialist. 
%     \label{fig:moe}}
% \end{figure}
%2% %%%%%%%%

%%%%%%%%
\begin{table}[t]
    \centering
    \scriptsize
    \begin{tabular}{@{}l l c S[round-mode=places,round-precision=1] S[round-mode=places,round-precision=1] S[round-mode=places,round-precision=1] S@{}}
        \toprule
        \multirow{2}{*}{Tasks}                     & \multirow{2}{*}{Datasets}              & \multirow{2}{*}{Metrics}     & {\tablecell{c}{OFA+\\(Specialist)}} & {\tablecell{c}{OFA+\\(Generalist)}} & {\tablecell{c}{OFA+\\(Generalist MoE)}} & {UnifiedIO} \\
        & & & { \numm186m $\times$ \num[round-mode=none]{15} } & { \numm270m } & { \numm455m } & { \numm776m } \\\midrule
        \multicolumn{5}{@{}l}{\textit{Text-only tasks}}                                                                                      \\
        Instruction Tuning                         & NaturalInstruction v2~\cite{natural_instruction_v2} & R-L $\uparrow$    
        & 30.49 & 26.97 & 27.74 & {-} \\
        Summarization & Gigaword~\cite{gigaword}            & R-L $\uparrow$     
        & 34.24 & 34.68 & 33.95 & {-}  \\\midrule
        \multicolumn{5}{@{}l}{\textit{Image-related tasks}}                                                 \\
        Classification & ILSVRC~\cite{imagenet} & Acc $\uparrow$         
        & 83.31 & 72.56 & 78.95 & {-}         \\
        Visual Entailment         & SNLI-VE~\cite{xie2019visual}               & Acc $\uparrow$        
        & 88.88 & 85.84 & 86.18 & 86.1      \\
        Captioning & COCO~\cite{chen15microsoft}  & CIDEr $\uparrow$     
        & 134.8 & 122.6 & 125.2 & 117.5       \\
        Visual Grounding          & RefCOCO~\cite{refcoco,kazemzadeh14referitgame}       & Acc@0.5 $\uparrow$     & 88.12              & 80.08             & 83.06       & {-}         \\
        VQA & VQA v2~\cite{goyal17vqav2}                & Acc $\uparrow$        & 77.28              & 68.24             & 71.72       & 67.8   \\
        Image Generation & COCO~\cite{chen15microsoft}                & CLIPSIM $\uparrow$        & {\tablenum[round-mode=none]{0.317}}             & {\tablenum[round-mode=none]{0.289}}             & {\tablenum[round-mode=none]{0.294}}       & {-}
        
        \\\midrule
        \multicolumn{5}{@{}l}{\textit{Audio-related tasks}}                                                                                  \\
        ASR                       & LibriSpeech~\cite{panayotov2015librispeech}           & WER $\downarrow$        & 7.5              & 8.5             & 8.1       & {-}         \\
        TTS                       & LJSpeech~\cite{ljspeech17}             & $\mathcal{L}$  $\downarrow$        &  {\tablenum[round-mode=none]{1.187}} &  {\tablenum[round-mode=none]{1.443}}  & {\tablenum[round-mode=none]{1.429}}   & {-}                                                         \\\midrule
        \multicolumn{5}{@{}l}{\textit{Video-related tasks}}                                                                                  \\
        Classification            & Kinetics400~\cite{imagenet}           & Acc $\uparrow$   & 74.30              & 64.58             & 69.47       & {-}         \\
        Captioning                & MSR-VTT~\cite{xu2016msr}               & CIDEr $\uparrow$      & 70.80              & 59.10             & 63.00       & {-}     \\
        VQA  & MSR-VTT QA~\cite{xu2017video} & Acc $\uparrow$ &  42.10              & 41.73             & 40.00       & {-}     
        \\\midrule
        % \multicolumn{5}{@{}l}{\textit{Motion-related tasks}}                                                                                 \\
        % Text to Motion            & AMASS/KIT/AIST++                    & loss          &                                                                    \\\midrule
        \multicolumn{5}{@{}l}{\textit{Structure-related tasks}}                                                                              \\
        Table-to-Text             & DART~\cite{DBLP:conf/naacl/NanRZRSHTVVKLIP21}                  & BLEU   $\uparrow$     & 51.24              & 50.86             & 50.88       & {-}         \\
        Text-to-SQL               & Spider~\cite{DBLP:conf/emnlp/YuZYYWLMLYRZR18}                & EM $\uparrow$ & 45.70                  & 39.20             & 40.50       & {-}         \\\midrule
        % \multicolumn{5}{@{}l}{\textit{Zero-shot tasks}}                                                                                      \\
        \multicolumn{3}{r}{Average (Performance)} & { \qty[round-mode=none]{100}{\percent} } & { \qty[round-mode=none]{91}{\percent} } & { \qty[round-mode=none]{95}{\percent} } & {-}       \\
        \multicolumn{3}{r}{Average (Model Size)} & { \qty[round-mode=none]{100}{\percent} } & { \qty[round-mode=none]{10}{\percent} } & { \qty[round-mode=none]{16}{\percent} } & {-}   
        \\
        \bottomrule
    \end{tabular}
    \caption{Experimental results on OFA+ (Specialist), OFA+ (Generalist), and OFA+ (Generalist MoE) according to the validation set (test-dev for VQA v2), with the exception of grounded image captioning and text-to-motion synthesis, which lack validation sets. 
    %All metric values are multiplied by \num{100}, including CLIPSIM and $\mathcal{L}$. 
    $\uparrow$ indicates the higher the better; $\downarrow$ indicate the lower the better. R-L stands for ROUGE-LCS~\cite{rougle}, Acc@0.5 stands for accuracy where $\text{IoU} \geq 0.5$ is considered correct~\cite{refcoco}, and EM stands for accuracy where an exact match is considered correct~\cite{DBLP:conf/emnlp/ZhongYK20}. %The tasks of grounded image captioning and text-to-motion synthesis 
    Results of Unified-IO are taken directly from \cite{lu2022unifiedio}. %, which rounds up the scores to the nearest tenth.
    }
    \label{tab:results}
\end{table}
%%%%%%%%

Both models are trained with 
%\zc{17 or 23, train on 23, test on 17?} 
\num{17} downstream tasks together involving data from \num{7} modalities. %of xx datasets.
For a detailed list of the task mixture, please refer to \cref{appx:settings}.
In evaluation, both models do not go through task-specific finetuning. 
We report results on the validation sets, except for VQA v2, where test-dev is commonly used as validation, and \num{2} tasks that do not come with validation set, \ie, grounded image captioning and text-to-motion synthesis. %, for which we sample generated examples.
% We select checkpoints that have relatively good performance on \num{15} diverse tasks according to the development set.
%Since TTS task requires more convergence steps, and other tasks may lead to over fitting, so here we briefly list the comparison of loss under the same steps.
%As Grounded Caption does not have evaluation subset and is %used in pretraining only, results on the development set of  %\num{16} datasets are reported.

For OFA+ (Generalist) and OFA+ (Generalist MoE), we use the AdamW~\cite{adamw} optimizer with $(\beta_1, \beta_2) = (\num{0.9}, \num{0.999})$ and $\epsilon = \num{1e{-8}}$ to train the model. 
We set the peak learning rate to $\num{3e{-4}}$, and apply a scheduler with linear decay with a warmup ratio of \num{0.01} to stabilize the learning. 
For more detailed settings, such as batch size for each task, please refer to \cref{tab:task parameter} in the appendix.
Both models are trained using \num{32} NVIDIA A100 80GB GPUs.

%For more detailed settings, such as batch size and learning %rate for each task, please refer to \cref{appx:settings}.

\subsection{Results and Findings}

% It should be noted that the results presented in this section is to demonstrate that \ofasysstyled{} as a library functions properly and the overall approach is scalable.
% Pursuing state-of-the-arts performance is neither the goal of this paper or in the scope of system implementation.

%It is always interesting to see what happens when a mixture of diverse tasks involving diverse modalities are trained together.

We report results on 15 tasks used in our training setup to evaluate a single jointly-trained model. 
In fact, to our knowledge, it is the first time that data from those 7 modalities are used together to train a single model, thanks to the support from \ofasysstyled{}.

In \cref{fig:learning-curves}, we show the learning curves of different tasks in the OFA+ (Generalist) model.
With the monitoring support in \ofasysstyled{}, we see that different tasks have divergent learning speed and may reach maximum performance at different steps, given a vanilla implementation of the universal model.
%It is also interesting to see that image classification and video classification have similar learning curves, suggesting task cooperation.
%However, relationships amongst different tasks require further investigation. 
%we further compare the learning curves of OFA+ (Generalist) and OFA+ (Generalist MoE) on representative tasks.
In \cref{fig:moe}, we further compare the learning curves of generalist models on different universal model structures.
As we can see, OFA+ (Generalist MoE), which is slot-wise sparsely-activated, converges faster and better than the other, regarding to the video and the audio tasks.
\ofasysstyled{} facilities such investigation of model structures by only modifying the universal model implementation or the model composition, while keeping the entire instruction set, \ie, tasks, fixed.
% In \cref{fig:moe}, we show the OFA+ (Generalist MoE) model in the current form is more performant that the OFA+ (Generalist) model and  converges faster and better than Generalist in different modes such as speech and video.

We compare the validation results between specialist models and generalist models in \cref{tab:results}. The OFA+ models trained on single tasks is referred to as OFA+ (Specialist).
%For reference, we also list results of the OFA+ models trained on single tasks, \ie, OFA+ (Specialist).
Note that due to the differences in model sizes, the  experimental results are not precisely comparable.
Comparing OFA+ (Generalist) and OFA+ (Generalist MoE), we can see that the latter performs much better on vision-related tasks.
The overall average score also understandably favors OFA+ (Generalist MoE).
%However, comparing to the specialist models, there is a long way to go for the multi-task trained models.
%However, limited to the size of the model, we have not started to discuss the problems on unseen tasks.
With only \qty{16}{\percent} parameters of the 15 task-finetuned specialist models, OFA+ (Generalist MoE) still achieves \qty{95}{\percent} of the specialists' performance in average, showcasing the potential of multi-modal task-scaling with \ofasysstyled{}.
%But it is still inspiring to see that OFA+ (Generalist MoE) achieves \qty{95}{\percent} of the performance with \qty{17}{\percent} of the parameters of 15 OFA+ (Specialist).

As the OFA+ (Generalist) and OFA+ (Generalist MoE) models have scaled the number of modalities to a new level, there is currently no precisely-comparable models from other studies. The most comparable method is Unified-IO~\cite{lu2022unifiedio} which conducts text and image tasks in a similar finetuning-free manner to us.
%Please also be aware that it is not a fair comparison: 
Compared to Unified-IO, generalist OFA+ models are trained on a different mixture of tasks with more modalities and fewer vision-language tasks, also with substantially fewer parameters. % (\numm270m for OFA+ and \numm455m for OFA+ (Generalist MoE) vs. \numm776m for Unified-IO Large).
%It is interesting to see 
The results demonstrate the generalist models trained by \ofasysstyled{} are better in terms of performance on the overlapping tasks.  
%Overall, we find that the task and modality mixture used in \ofasysstyled{} induces a better performance than UnifiedIO.
% The reason may come from the fact that data from more modalities are generally helpful for model training.

In all, in application to OFA+, \ofasysstyled{} demonstrates evidently its functionality, scalability, and flexibility. 
The functionality is validated by training and conducting inference using a number of models on diverse tasks in single-task or multi-task settings.
The full-fledged support allows meaningful exploration of generalist models beyond the language-vision or language-speech settings.
The scalability lies in the composition of \num{23} diverse tasks over \num{7} modalities with a consistent declarative interface.
They can theoretically be used to train a single model all together by reusing instructions even without writing new code.
% One can create a new task without implementing new methods in just a single line of code.
The flexibility is shown by the two versions of OFA+ (Generalist): the computation pipeline is decoupled reasonably enough for one to focus research on a part of the whole system.
For example, one can focus on model structures \wrt modality by modifying only the universal model, without worrying breaking data processing pipelines.

% Finally, for the tasks without validation set, we show sampled outputs in \cref{fig:examples}.

% \placeholder{examples and showcases, possibly with zero-shot and interesting new tasks}
% %%%%%%%%
% \begin{figure}[t]
%     \centering
%     \fbox{\rule{0pt}{18\baselineskip} \rule{0.9\linewidth}{0pt}}
%     \caption{Inference examples on the tasks without validation sets.}
%     \label{fig:examples}
% \end{figure}
% %%%%%%%%

\section{Conclusion and Future Work}
\label{sec:conclusion}

As generalist models attract increasing interests, the lack of designated system and library for multi-modal multi-task stands out as an obstacle in the path for rapid growth.
\ofasysstyled{} is developed to match the need in multi-modal multi-task learning of extreme modality and task scaling.
% To scale in terms of tasks and modalities, \ofasysstyled{} proposes to think beyond the conventional concepts and provides an interface that responds better to the experimental needs.
With \ofasysstyled{}, it is easy to (a) rapidly introduce new multi-modal tasks by defining a declarative instruction in a single line of code, and (b) easily introduce, reuse, and customize modality-specific components.
The functionality is realized by a carefully-designed library structure that decouples the complexity of multi-modal multi-task learning into a hierarchy of components.
% The instruction defines a job which is the realization of a task on a specific dataset that is scheduled by \ofasysstyled{}.
% Templates and slots decouple the tasks and modalities, enabling flexible composition of pre-defined components.
We train a series of models named OFA+, using \ofasysstyled{}, and show that it is achievable for a generalist model to understand data from more modalities and perform promisingly.
In all, we hope \ofasysstyled{} would push forward the research in multi-modal multi-task learning and facilitate the construction of generalist models that are even more general.

In the current development phase, we focus on the functionality of the instruction interface and the system completeness and correctness, including but not limited to offering as many as possible slot types and task examples, and a smooth training and inference experience.
In addition, although \ofasysstyled{} is designed with generalist models in mind, the design of the declarative interface and the hierarchy of reusable components also streamline machine learning research on single tasks.
However, there are still many areas where \ofasysstyled{} can be improved.
For example, as related research progresses rapidly, \eg, the development of diffusion-based models, we actively explore appropriate ways to unify them with the system but have to leave the accommodation of U-Net model structures to the future.
The emergence of new learning paradigms, \eg, multi-modal in-context learning through interleaved image-text data, also presents challenges that are currently being addressed.
Nonetheless, we feel that in the current state \ofasysstyled{} can already serve as a concrete platform for future algorithm research and we are more than happy to share our efforts with the community.

% \subsubsection*{Acknowledgement}

% We thank the author of fairseq and taming transformers ...

% \newpage
\appendix

\section{Advanced Usage}
\label{sec:interface}

In \cref{sec:usage}, we illustrate the high-level basic usage of \ofasysstyled{}.
As we can see, a multi-modal task can be defined in a single line of code, which facilities task scaling via comprehensive default implementations.
However, it is often one's desire to control the system with finer-granularity.
\ofasysstyled{} provides two ways to achieve this goal, the YAML configuration that introduces structured parameters to instructions and a conventional imperative interface to programmatically define common procedures in computation pipelines.
The first option is discussed in \cref{sec:training,sec:inference}; and the second option is summarized in \cref{sec:extend}.

\subsection{Multi-Task Training with Instructions}
\label{sec:training}

Although it is straight-forward to use instructions directly in code, to achieve better reusability in training, \ofasysstyled{} can be also configured via configuration files.
In experimentation, the configurations of different tasks, models, and trainers can be easily combined to build a new multi-modal multi-task learning system.
\ofasysstyled{} organizes the training configuration files consistent with the routine in \cref{sec:usage}.
The configuration files adopt the YAML format\footnote{\url{https://github.com/yaml/yaml-spec}}, which is human friendly and suitable for structured data.

\subsubsection{Tasks}

For tasks, the configuration file should at least contain the instruction section and the dataset section.
An example file for the captioning task is given in the following:
\begin{yamlcode}
# caption.yaml
instruction:
  - '[IMAGE:img] please use a short line to describe the image. -> [TEXT:cap]'
  - '[IMAGE:img] what does the image describe? -> [TEXT:cap]'
  
dataset:
  train_data: coco_2014/train
  valid_data: coco_2014/valid
  update_freq: 2
  micro_batch_size: 8

preprocess:
  image:
    patch_image_size: 480
  text:
    max_src_length: 128
    max_tgt_length: 20

criterion:
  label_smoothed_cross_entropy:
    label_smoothing: 0.1

evaluation:
  metrics:
    cider: 
      target_field: cap
  generator_args: '{"beam":5,"max_len_b":16,"no_repeat_ngram_size":3}'
\end{yamlcode}
This configuration file presents a more structured view of the task pipeline components and showcases several advanced usage of \ofasysstyled{}, compared to \cref{sec:usage}, where all default values are used.

First, it can be seen that multiple instructions can be specified for a dataset.
For each example in the dataset, \ofasysstyled{} currently randomly samples an instruction according to a uniform distribution.
However, it is also required that the instructions should be \concept{collation-compatible}, which means they should have the same number of slots and the sequence of slot types should be exactly the same.

Second, the dataset section specifies the paths of data used in training. 
As there are complications in declaratively describing how to construct the dataset, the data paths in configuration files should be either a HuggingFace Dataset spec, or a URI to a Tab-separated Values (tsv) file.
The URI can be a local file path (optionally starting with file://) or a remote file path (where the protocol is a must).
The tsv file should have a header and is parsed and processed as a \func{torch.utils.data.IterableDataset}.
The dataset section also includes hyper-parameters in training the task, such as \param{micro\_batch\_size} and \param{update\_freq}.
The former is the per GPU batch size, and the latter is for gradient accumulation.
Although they are related to optimization, they are included to streamline task composition in multi-task training.
For more details, please refer to the documentation of \ofasysstyled{}.

Finally, there are optional configurations for task-specific pipelines. 
Slots of the same type share common preprocessor, postprocessor, and adapter configurations by default.
These can be adjusted in the preprocess and the adapter sections.
For example, the \param{max\_src\_length} sets the maximum length of every encoder \texttt{TEXT} slot to \num{128}.
The criterion used to obtain the loss can be automatically deducted from the instructions, which is the criterion based on cross entropy most of the time.
A common usage is to add label smoothing normalization to the cross entropy, which is shown in the preceding example.
The evaluation section tells how the performance of the task on this dataset can be evaluated.
The \param{generator\_args} is for the generator in inference to obtain the prediction and the metric section can include multiple metrics to evaluate the prediction against the ground truth in \param{target\_field}.

For your information, the name in each section, \eg, \func{image} and \func{text} in preprocess, \func{label\_smoothed\_cross\_entropy} in criterion, and \func{cider} in metrics, is associated with the implementation via a hierarchical configuration registry.
For a complete list of supported variants, please refer to the \ofasysstyled{} documentation.

The configuration file can be used to initialize a task with a dataset in one step:
\begin{pythoncode}
from ofasys import Task
task = Task.from_yaml("caption.yaml")
\end{pythoncode}
There are \num{23} readily-available task configuration files at the time of writing this manuscript.
Please see \cref{appx:tasks} for reference.

\subsubsection{Universal Models}

For universal models, the default configuration covers all aspects and no user input is required.
The following file customizes the model implementation for all tasks:
\begin{yamlcode}
# model.yaml
model:
  arch: large
  use_fused: true
  freeze_encoder_embedding: true
  freeze_decoder_embedding: true
  encoder_drop_path_rate: 0.2
  decoder_drop_path_rate: 0.2
  layernorm_position: true
\end{yamlcode}
The \param{arch} specifies a set of configuration for model architecture, \eg, embedding dimension and number of layers.
The rest parameters further finetune the structure.
For example, \param{use\_fused} makes \ofasysstyled{} use fused CUDA kernel implementation for certain modules to improve computation performance on GPU devides.
Encoder and decoder computation in training is adjusted via \param{freeze\_*\_embedding} and \param{*\_drop\_path\_rate}, which indicates the adapters in the models are not optimized and a drop path normalization is used for both the encoder and the decoder.
For complete configurable options, please also refer to the \ofasysstyled{} documentation.

The model can be initialized from files similarly to tasks:
\begin{pythoncode}
from ofasys import GeneralistModel
model = GeneralistModel.from_yaml("model.yaml")
\end{pythoncode}
It should be well known that every tasks would share the same universal model structure.

\subsubsection{Trainer}

The trainer implements logistics in multi-task training.
An example configuration file is given in the following:
\begin{yamlcode}
# trainer_conf.yaml
common:
    fp16: true
    fp16_scale_window: 512
    log_interval: 10

distributed_training:
    find_unused_parameters: true

optimization:
    max_update: 10000
    clip_norm: 1.0
    lr: [1e-5]
    sentence_avg: false

optimizer:
    adam_betas: [0.9, 0.999]
    adam_eps: 1e-08
    weight_decay: 0.01

lr_scheduler:
    warmup_ratio: 0.06
\end{yamlcode}
The common section includes configuration for overall computation, profiler and monitor.
The rest sections configure parameter optimization and others.

Considering memory efficiency, all tasks share a common optimizer and thus the same learning rate schedule, as most optimizers for multi-modal training estimate gradient momentums which diverge with different learning rates.
Currently, the default optimizer is AdamW~\cite{adamw} and the default learning rate scheduler is based on linear decay with warmup~\cite{liu19roberta}.
The options are being actively extended.

The trainer can be initialized as
\begin{pythoncode}
from ofasys import Trainer
trainer = Trainer.from_yaml("trainer_conf.yaml")
\end{pythoncode}

\ofasysstyled{} also includes a launch script to conveniently combine the YAML configuration files in training. 
The launch script also wraps distributed training, which is highly environment-dependent. 
For related usage, please refer to the \ofasysstyled{} documentation. % for the usage.

\subsection{Inference with Instruction}
\label{sec:inference}

As tasks are expressed as instructions, \ofasysstyled{} supports task-agnostic inference using the trained models, which includes conducting novel multi-modal tasks.
% Being a unified encoder-decoder model, the decoder is tasked to generate the output.
% The inference interface is streamlined to reflect these characteristics.

As prerequisite, inference needs a pretrained model checkpoint. 
Especially the modality-related structures need training, if the modality is desired in inference.

%OFA-Sys can do task-free multi-modal inference on just one checkpoint through the instruction alone. First, we load a pretrained checkpoint:
\begin{pythoncode}
from ofasys import OFASys
model = OFASys.from_pretrained('my-awesome-checkpoint.pt')
# Inference for image captioning
instruction = '[IMAGE:img] what does the image describe? -> [TEXT:cap]'
data = {'img': 'image_1.jpg'}
output = model.inference(instruction, data=data)
print(output.text) # the caption
\end{pythoncode}
As we can see, one only need to pass the instruction and the data for inference.
The model checkpoints saved by \ofasysstyled{} keep all the training configurations so there is no need to include them again. 
% As we can see, the task is no longer required in inference.
Although the instruction is preferably similar to the ones used in training, there is no strict enforcement from the library side.
The results can be retrieved with the slot type being the key.
%For more examples, please refer to the \cref{sec:more-inference}.

\ofasysstyled{} also provides convenience methods to save \texttt{BOX} and \texttt{MOTION} output data. 
More inference examples are shown below:

Image Grounding (Auto-Regressive Token Generator):
\begin{pythoncode}
instruction = '[IMAGE:img] which region does the text " [TEXT:cap] " describe? -> [BOX:region_coord]'
data = {'img': 'image_1.jpg', 'cap': 'hand'}
output = model.inference(template, data=data)
output.save_box('output_with_bbox.jpg')
\end{pythoncode}

Automatic Speech Recognition (Auto-Regressive Token Generator):
\begin{pythoncode}
instruction = '[AUDIO:wav] what is the text corresponding to the voice? -> [TEXT:txt]'
data = {'wav': 'audio.flac'}
output = model.inference(template, data=data)
print(output.text)
# nor is mister quilters manner less interesting 
# than his matter
\end{pythoncode}

Text-to-Motion Synthesis (Diffusion Generator):
\begin{pythoncode}
instruction = 'motion capture: [TEXT:text] -> [MOTION:bvh_frames]'
data = {'text': 'run then jump'}
output = model.inference(template, data=data)
output.save_as_gif('run_jump.gif')
\end{pythoncode}

\subsection{Extensibility and Interoperability}

\label{sec:extend}

Apart from the declarative user interface, \ofasysstyled{} provides an imperative user interface aiming for extensibility.
Users can easily build experiments with their new ideas by extending \concept{base classes}.
To integrate the custom components into the system, one needs to use the decorator \func{@register\_config}.
% Configurations of the new components can be registered into the system by using the decorator ``@register\_config''.
%After registration, users can specify and use their newly added modules in the instruction just like the system pre-defined modules.
For detailed how-to tutorials, please refer to the \ofasysstyled{} documentation.

\paragraph{Adding Task-Specific Data Processing Methods}

% Registering a new task is probably the most common custom development for users.
Although the modality presets can cover many of the data processing needs, one can further customize the data processing operations by extending the base class \func{BaseTask} with their own preprocessing/postprocessing logic.
Those functions are called before/after the system preprocessor/postprocessor.

% One can simply write a \concept{instruction} to define an execution logic specifying which parts of the model should be involved in dealing with certain input-output mapping.
% Besides, we provide a base class \concept{OFATask} which contains several default helpers for loading data, training, and generating.
% For users have to deal with special operations required for some datasets or tasks, they can extend this base class and add their own logic.

\paragraph{Adding New Modality/Slot Type}

For adding new modality, one needs to implement a preprocessor and an adapter at the minimum.
Preprocessors should extend the \func{SafeBasePreprocess} class, which contains a sanity check for the inputs. 
% The required methods are \concept{map}, \concept{group\_map} and \concept{collate}.
Adapters should extend the \func{BaseAdapter} class, which extends \func{torch.nn.Module}.
The main difference between preprocessors/postprocessors and adapters from the library side is that adapters can be trained together with the model and their parameters must be saved in the model checkpoints.

\section{Modality Suport}
\label{appx:slots}

\ofasysstyled{} provides \num{7} presets for \texttt{TEXT}, \texttt{IMAGE}, \texttt{VIDEO}, \texttt{AUDIO}, \texttt{BOX}, \texttt{STRUCT}, and \texttt{MOTION} modality slots.
Please note that the slots are categorized by its raw input data modality, not the transformed inner data modality.

\subsection{\texttt{TEXT}}

% type/input data type introduction
\texttt{TEXT} is the most common slot type, as many other forms of structural data can be transformed into/from text, \eg, category labels and table schemas. 
% preprocessor
The default preprocessor conducts (sub-)tokenization using GPT2BPE~\cite{radford19gpt2} and can apply masking for masked language modeling.
There is another preprocessor that can transform text to phone used in audio tasks.
% postprocessor
The default postprocessor decodes tokens into text losslessly, as GPT2BPE is whitespace-aware.
% adapter
The adapter encodes and decodes the token or phone sequence using text embeddings or phone embeddings.
As adapters of the same type share parameters, for sequence-to-sequence learning, it means source input embedding, target input embedding, and target output embedding are all shared.
% criterion
% TEXT slotserve as both E-Slot and D-Slot.
The available training objectives include MLE, SCST, and InfoNCE.
% generator
As tokens are discrete, the currently supported generator is the auto-regressive token generator.

\subsection{\texttt{IMAGE}}
\texttt{IMAGE} is another common slot type containing visual spatial data in different formats. 
% preprocessor
The default preprocessor provides a simple series of transformations, such as resizing and normalization, to convert raw image data into the tensor format.
We also provide a more complex preprocessor, which contains several data augmentation steps.
% postprocessor
\texttt{IMAGE} slot uses postprocessor only in image generation tasks and we use VQ-GAN~\cite{esser21taming} to convert discrete code sequence back to images.
% adapter
% Image slot need different adapters when serve as E-Slot and D-slot.
We provide ResNet~\cite{resnet} and ViT~\cite{vit} as image adapters for encoder slots.
For the decoder slot in image generation task, we use VQ-GAN to encode images into discrete code sequences.
% criterion
The available training objective is MLE using cross-entropy loss for image codes.
% generator
Similar to \texttt{TEXT}, we use auto-regressive token generator to generate the discrete image code sequence.

\subsection{\texttt{VIDEO}}
% type/input data type introduction
\texttt{VIDEO} is a slot type for consecutive image frames extracted from a video, \ie, tempo-spatial data.
% preprocessor
The default preprocessor can decode the video and extract frames using a specified downsampling rate. 
It also supports data augmentations, following MViT \cite{fan2021multiscale} and PySlowFast \cite{fan2020pyslowfast}.
% adapter
Currently, \texttt{VIDEO} can be only used as encoder slots. 
The adapter builds upon and reuses pretrained weights of the \texttt{IMAGE} adapter (ViT or ResNet) to encode each frame, and concatenate them to a sequence of representation vectors.
% Additional frame-level positional encodings are added to the original image positional encoding. 
% All weights from the Image adapter are reused.
% criterion
The available training objectives include MLE and SCST.
% generator
The \texttt{VIDEO} slot currently does not has a postprocessor or a generator. 
However, it is possible to cast the video generation task as image generation task: reuse the auto-regressive discrete generator to generate a single video frame and append the frame to the source; by repeating this step, we can generate a video.

\subsection{\texttt{AUDIO}}
% type/input data type introduction
\texttt{AUDIO} is a slot type for sound features in the time and frequency domains. %, including raw audio waveform and log Mel-filterbank feature.
% preprocessor
The default preprocessor can extract the log Mel-filterbank features from raw audio waveform. 
For robust speech pprocessing, techniques including (a) volume normalization and (b) cepstral mean and variance normalization (CMVN)~\cite{prasad2013improved} can be applied. 
Besides, specAugment~\cite{park2019specaugment} and speed perturbation~\cite{ko2017study} are employed for data augmentation.
% postprocessor
\texttt{AUDIO} uses a postprocessor only in generation tasks, where we transform the predicted mel spectrograms to waveform via a vocoder, \ie, HiFi-GAN~\cite{kong2020hifi}.
% adapter
Currently, the adapter can operate on log Mel-filterbank features. 
In the encoder slots, the \texttt{AUDIO} input adapter consists of a CNN for downsampling and a transformer network for contextutal representation learning.
In the decoder slots, it employs a fully-connected network as the input adapter and a CNN as the output adapter following Tacotron2~\cite{shen2018natural}.
% criterion
The available training objectives include MSE of mel spectrograms and cross-entropy loss on the stop probability for speech synthesis.
% generator
We utilize an auto-regressive feature generator for speech generation since fbank features are continuous.
%the Audio slots differ from Text and Image slots in that they are continuous-valued sequences.

\subsection{\texttt{BOX}}
% type/input data type introduction
\texttt{BOX} is a slot type for processing bounding boxes in region-like tasks (\eg, object detection, grounded captioning, and visual grounding). 
% preprocessor
The default preprocessor can quantize the continuous corner coordinates (top-left and bottom-right) of the bounding box to discrete box tokens $\langle x_1,y_1,x_2,y_2 \rangle$ \cite{pix2seq}.
The preprocessor allows the user to set the range of discrete values to control the granularity.
% postprocessor
\texttt{BOX} implements a postprocessor to recover the continuous corner coordinates from box tokens, so users can visualize the bounding boxes on the image.
% adapter
Since the coordinates have been converted to discrete box tokens, the \texttt{TEXT} adapter can be also used to encode the tokens into a sequence of embeddings.
% criterion
The available training objective is MLE.
% generator
As box tokens are discrete, the currently-supported generator is the auto-regressive token generator.

\subsection{\texttt{STRUCT}}
% @麻油
% type/input data type introduction
Structural data, such as databases, tables, grids, graphs, and trees, is widely used in many areas, \eg, knowledge graph and protein structure. 
\texttt{STRUCT} is the slot for structural data, and currently, it supports table and database data. 
% preprocessor
Inspired by UnifiedSKG~\cite{DBLP:journals/corr/abs-2201-05966}, the default preprocessor transforms the structural data into sequential text data. %and inherit the function of text slot for the next processings. 
For tables of small sizes, the preprocessor flattens the whole table into a sequence, using ``$:$'' to distinct each column and ``$|$'' to distinct each row. 
For tables of large sizes or database schemas, the preprocessor only extracts the schemas information with a few mentioned row names from the instructions. 
% postprocessor
Since \texttt{STRUCT} is only used as generation target in sudoku, we implement a text-to-sudoku postprocessor.
% The \texttt{STRUCT} slot currently does not have a postprocessor, as it is not used for any structural data generation tasks.
% adapter
Since the structural data are converted to text, the \texttt{TEXT} adapter can be used to encode the text tokens into a sequence of embeddings.
% criterion
The available training objective is MLE. %, as the \texttt{TEXT} adapters is used.
% generator
As the transformed text tokens are discrete, the currently supported generator is the auto-regressive token generator.

\subsection{\texttt{MOTION}}
\label{appx:motion}
Human motion, the \texttt{MOTION} modality, is common in 3D character animation, robotics, and human behavior understanding.
%A clip of human motion consists of multiple frames, where each frame describes a humanoid actor's pose at that moment.
%The key information describing a pose is (a) the current position of the root joint, typically the hip, in the 3D coordinate system, and (b) how all the joints of the human body should rotate in order to form the pose, \ie., the rotation matrices of the joints.
%
We implement the \concept{motion\_6d} preprocessor for motion data.
It reads a BVH file, which is a motion capture data format commonly used by the industry to describe a clip of motion, and converts the BVH file into a floating-point number (float) array of shape $[n, 6 + 6m]$, where $n$ is the number of frames and $m$ is the number of joints. %[\#frames, 6+6$\times$\#joints]. 
Each frame describes the pose at that moment, which includes \num{3} floats describing the root joint's 3D position, another \num{3} floats for the body's velocity, and $6m$ floats corresponding to all the joints' rotations.
Note that we convert each $3\times 3$ rotation matrix into its 6D vector representation~\cite{6d-rep} to ease learning.
The \concept{motion\_6d} adapter employs linear projection to align the dimension of the data and the universal model.
By default, we use a criterion and a generator suitable for measuring the generation error of a continuous signal to handle motion data, i.e., the denoising diffusion probabilistic modeling (DDPM)~\cite{ho-ddpm} loss and its corresponding generator.
DDPM requires the adapter to handle the step information of DDPM, where the step information can also be understood as the strength of the noise for data corruption.
The adapter incorporates this step information by adding a step embedding to the token embeddings, similar to how position embeddings are usually implemented.
%We also include an optional tokenizer, similar to VQ-GAN, for quantizing motion in the adapter, if one prefers a discrete loss such as the causal language modeling objective.

% An adapter, also named \concept{motion\_6d}, is implemented to accompany the preprocessor.
% A clip of motion is viewed as a sequence with length equal to the number of frames.
% The adapter uses a linear layer to project the data into a latent space such that the dimension of each frame's embedding is consistent with the one used by the universal compute engine.
% The output of the universal compute engine is also projected back to the original data dimension linearly by the adapter.

\section{Task Examples}
\label{appx:tasks}

\ofasysstyled{} currently includes \num{23} example tasks with default configurations.
In the following, we summarize the default settings for those tasks.

\subsection{Text-Only Tasks}

\subsubsection{Text Understanding (GLUE)}

\paragraph{Task Introduction:}
% ofasys无关的task介绍，包括task本身对原task/dataset的转换。
GLUE~\cite{wang19glue} is a benchmark for text understanding, which casts multiple datasets into a unified sentence classification form.
The tasks/datasets include the Corpus of Linguistic Acceptability (CoLA)~\cite{warstadt19cola}, the Stanford Sentiment Treebank (SST-2)~\cite{socher13recursive}, Microsoft Research Paraphrase Corpus (MRPC)~\cite{dolan05mrpc}, Semantic Textual Similarity Benchmark (STS-B)~\cite{cer17sts}, Quora Question Pairs (QQP)~\cite{qqp}, MultiNLI (MNLI)~\cite{williams18mnli,bowman15snli}, Question NLI (QNLI)~\cite{rajpurkar16squad,white17inference,demszky18transforming}, Recognizing Textual Entailment (RTE)~\cite{dagan05rte1,barhaim06rte2,giampiccolo07rte3,bentivogli09rte5}, and Winograd NLI (WNLI)~\cite{levesque11winograd}.
A majority of the original datasets are cast as natural lanugage inference (NLI) tasks (identifying entailment, neutral, contrast relationships) or binary classification tasks (yes/no).
The evaluation metric is Matthew's Correlation Coefficient~\cite{matthews1975} for CoLA, Pearson's and Spearman's Correlation Coefficient for STS-B, and accuracy for the rest.
The overall score for this task in evaluation is commonly the arithmetic average over the \num{8} tasks without WNLI~\cite{devlin2018bert}.
Please note that currently \ofasysstyled{} is not able to support regression tasks in GLUE, \ie, STS-B, which will be addressed soon, as there are no technical obstacles.

\paragraph{Default Instruction:} 
% ofasys中相关的可以通过intruction表达的内容，包括ofasys本身的任务变形、预处理、适配器、生成器、特殊处理等。如果有相当多的算法创新，请尽可能简要的说明，不必具体到代码实现。
The default instructions for this task are as follows:

\noindent
CoLA: 
\begin{inst}
is the text "[TEXT:s]" grammatically correct?  -> is the text "[TEXT:s,no_loss]" grammatically correct? [TEXT:label,closed_set]
\end{inst}
SST-2:
\begin{inst}
is the sentiment of text "[TEXT:s]" positive or negative? -> is the sentiment of text "[TEXT:s,no_loss]" positive or negative? [TEXT:label,closed_set]
\end{inst}
QQP:
\begin{inst}
is question "[TEXT:q1]" and question "[TEXT:q2]" equivalent? -> is question "[TEXT:q1,no_loss]" and question "[TEXT:q2,no_loss]" equivalent? [TEXT:label,closed_set]
\end{inst}
MRPC:
\begin{inst}
does text1 "[TEXT:s1]" and text2 "[TEXT:s2]" have the same semantics? -> does text1 "[TEXT:s1,no_loss]" and text2 "[TEXT:s2,no_loss]" have the same semantics? [TEXT:label,closed_set]
\end{inst}
MNLI:
\begin{inst}
can text1 [TEXT:s1] imply text2 [TEXT:s2]? -> can text1 [TEXT:s1,no_loss] imply text2 [TEXT:s2,no_loss]? [TEXT:label,closed_set]
\end{inst}
QNLI:
\begin{inst}
does "[TEXT:s]" contain the answer to question "[TEXT:q]"? -> does "[TEXT:s,no_loss]" contain the answer to question "[TEXT:q,no_loss]"? [TEXT:label,closed_set]
\end{inst}
RTE:
\begin{inst}
can text1 "[TEXT:s1]" imply text2 "[TEXT:s2]"? -> can text1 "[TEXT:s1,no_loss]" imply text2 "[TEXT:s2,no_loss]"? [TEXT:label,closed_set]
\end{inst}
The input sentence(s) are repeated for better performance, as observed in \cite{wang22ofa}.
\param{no\_loss} indicates the repeated input sentences do not have targets in computing the loss.
%The text are preprocessed using GLUE-specific procedures and the common text preprocessor.
%The adapter is the text token embedding.
The generator in inference is the sequence generator with the token space constrained to a dataset-specific space via \param{closed\_set}.
%The post-processor maps the generated strings to the original label.

% \paragraph{Training:}
% % ofasys中默认的训练相关优化参数。
% For single task training, \ofasysstyled{} provides a set of default hyper-parameters suited to the OFA-Base checkpoint. 
% The batch size is \num{32}, the learning rate is \num{5e{-5}}, the number of maximum updates is equivalent to \num{10} epochs, and the warmup steps are \qty{10}{\percent} of the total updates.
% For normalization in training, the dropout path of rate \num{0.1} is used.

% \paragraph{Inference:}
% For generator, greedy decoding is used.

\subsubsection{Text Summarization (Gigaword)}

\paragraph{Task Introduction:}

Text summarization is a natural language generation task, where the model should produce a piece of concise text that covers the main points of the given text.
\ofasysstyled{} currently evaluates text summarization on Gigaword, following related work~\cite{gigaword}.
Gigaword for summarization is a naturally-annotated dataset consisting of news articles, where the first sentence of the article is regarded as the summary for the rest of the first paragraph.
The evaluation metric is ROUGE~\cite{rougle}.
Especially, we report ROUGE-L (R-L) in this paper.

\paragraph{Default Instruction:} The default instruction for this task is as follows:
\begin{inst}
what is the summary of article " [TEXT:src] "? -> [TEXT:tgt,noise_ratio=0.2]
\end{inst}
The \param{noise\_ratio} is passed to the preprocessor, which randomly replaces tokens with the specified rate in the target output.

% \paragraph{Training:} 
% For single task training, the batch size is \num{512}, the learning rate is \num{1e{-4}}, and the number of maximum training epochs is \num{6}. 
% The normalization methods include label smoothing of \num{0.1} and drop path of \num{0.1} on both the encoder and the decoder.

% \paragraph{Inference:}
% For the generator, we set beam size to \num{6}, length penalty to \num{0.7}, the number of generated tokens is limited to fewer than $L + 32$, where $L$ is the length of the original text, and no\_repeat\_ngram\_size is set to \num{3}.

\subsubsection{Natural-Instructions v2} 

\paragraph{Task Introduction:} 

Natural-Instructions v2 \cite{natural_instruction_v2} is a benchmark of over \num{1600} diverse language tasks which evaluates generalization across language tasks by leveraging their language instructions. 
It covers over \num{70} distinct task types, such as tagging, in-filling and rewriting. These tasks are collected with contributions of NLP practitioners in the community and through an iterative peer review process to ensure their quality.
Natural-Instructions v2 consists of a variety of language tasks and instructions that describe them in plain language. 
Each sample contains four fields. 
Instruction defines a given task in plain language. 
This involves a complete definition of how an input text (\eg, a sentence or a document) is expected to be mapped to an output text. 
Examples are samples of inputs and correct or wrong outputs to them, along with a short explanation for each.
On average, each sample contains \num{2.8} positive and \num{2.4} negative examples. 
Src and tgt are a large collection of input-output pairs for each task. 
Since this benchmark contains a large collection of tasks, we split the tasks into two subsets: 
one subset for evaluation and the remaining ones which can be used for supervision. 
For evaluation tasks, specifically, we fix a manually-selected collection of \num{12} categories that represent \num{154} tasks. 
We report ROUGE-L \cite{rougle} for aggregated performance results across a variety of tasks which is a soft string overlap metric that can be applied to a wide range of text generation tasks. 

\paragraph{Default Instruction:} The default instruction for this task is as follows:
\begin{inst}
[TEXT:instruction] [TEXT:examples] [TEXT:src] -> [TEXT:tgt,max_length=128]
\end{inst}
The maximum input length is set to \num{1024} and the output \param{max\_length} is set to \num{128}.

% \paragraph{Training:} 
% For single task training, the batch size of \num{512} and a learning rate of \num{1e{-5}} is applied..

\subsubsection{Text Infilling}

\paragraph{Task introduction:}
Self-supervised learning in natural language processing can belong to various forms.
In \ofasysstyled{}, we give an example for the most basic form of self-supervised learning for sequence-to-sequence models, that is, text-infilling, in a similar manner to BART~\cite{lewis20bart}.
For this task, the tokens in the input text is randomly replaced with a special mask token, and the model should recover the input text as the output.
In our experiments, the data is obtained from English Wikipedia.
This task is designed for pretraining only.

\paragraph{Default instruction:}
The default instruction for this task is as follows:
\begin{inst}
what is the complete text of "[TEXT:text,mask_ratio=0.3]"? -> [TEXT:text]
\end{inst}
The \param{mask\_ratio} is passed to the preprocessor, which randomly replaces input tokens with a special mask token.
% \paragraph{Optimization}

\subsection{Image-Related Tasks}

\subsubsection{Image Classification (ILSVRC-2012)}
\paragraph{Task Introduction:} 

Image classification task requires the model to predict the correct category for the input image. 
We evaluate our model on the ILSVRC-2012 ImageNet dataset~\cite{imagenet}. 
The dataset contains \numk1k image categories and around \numm1.3m images. 
Each image is manually annotated with one category label among the \numk1k candidates. 
We report the top-\num{1} accuracy on the test set of \numk50k images.

\paragraph{Default Instruction:} The default instruction for this task is as follows:

\begin{inst}
[IMAGE:image,preprocess=imagenet] what does the image describe? -> [TEXT:label_name,closed_set]
\end{inst}
% For this task, the text instruction is fixed. 
The input image resolution is set to \num{480}.
For the input image, we specify a special ImageNet preprocessor to activate the dataset-specific image augmentation on training samples. 
Specifically, following~\cite{beit}, we employ the same random resize cropping, random flipping, RandAug~\cite{rand_augment} and random erasing~\cite{random_erase} as data augmentation strategies on the training images. 
For the decoder slot, we add the specification \param{closed\_set} to constrain the text output into the $1$K candidate category names.

% \paragraph{Training:} 
% For single task training, the model is trained for \num{32} epochs, with a batch size of \num{256} and a learning rate of \num{5e{-5}} with \numk5k warmup steps. 
% The ratio of label smoothing is $0.1$. 
% The stochastic depth is employed in encoder and decoder with rate of $0.1$. 

\subsubsection{Image Captioning (COCO Captions)}
\paragraph{Task Introduction:}
Image captioning requires the model to generate a descriptive text for an image. 
We evaluate the multi-modal generation capability of \ofasysstyled{} on the most widely used COCO Caption dataset~\cite{chen15microsoft}. Following previous works~\cite{bottomup,wang22ofa}, We report CIDEr~\cite{cider} scores on the Karparthy test split~\cite{karpathy}.

\paragraph{Default Instruction:} 
The default instruction for this task is as follows:
\begin{inst}
[IMAGE:img] what does the image describe? -> [TEXT:cap]
\end{inst}

% \paragraph{Training:} 
% For finetuning, we first train the model with cross-entropy
% loss for \num{2} epochs with a batch size of \num{128} and a learning rate of \num{1e{-5}}, and label smoothing is set to \num{0.1}. 
% We then finetune the model with CIDEr optimization for \num{5} epochs with a batch size of \num{64} and a learning rate of \num{1e{-5}}, and disable dropout and stochastic depth.

\subsubsection{Visual Entailment (SNLI-VE)}

\paragraph{Task Introduction:}

Visual entailment (VE)~\cite{xie2019visual} is similar to textual entailment. It changes the premise from the text to the image, and judges whether the images matches the sentence. SNLI-VE is a data set of VE tasks which gives images, image captions and premises, and requires the model to judge the relationship between images and premises, and gives one of three outcomes: entailment, neutral, and contradiction.

\paragraph{Default Instruction:} 
The default instruction for this task is as follows:
\begin{inst}
[IMAGE:img] can image and text1 "[TEXT:cap]" imply text2 "[TEXT:hyp]"? -> can image and text1 "[TEXT:cap,no_loss]" imply text2 "[TEXT:hyp,no_loss]"? [TEXT:label,closed_set]
\end{inst}
The pattern is similar to MNLI task, where the encoder text is repeated in the decoder.
The input image resolution is resized to \numproduct{480 x 480} by default.

% \paragraph{Training:} 
% The model is trained for \num{6} epochs with a learning rate of \num{2e{-5}} and a batch size of \num{256}, and the warmup steps are \num{10} of the total updates. 

\subsubsection{Visual Question Answering (VQA v2)}

\paragraph{Task Introduction:}
Visual question answering (VQA) requires the model to answer questions based on the information of the given image~\cite{vqa, goyal17vqav2}. 
We finetune our pretrained model on the dataset VQA v2~\cite{goyal17vqav2}. 
We evaluate the performance by calculating accuracy. 

\paragraph{Default Instruction:} 

The default instruction for this task is as follows:
\begin{inst}
[IMAGE:image] [TEXT:question] -> [TEXT:answer,closed_set]
\end{inst}
% The model should predict the correct answer from a closed set of answers based on a given image and question. 
The resolution of the input images is \numproduct{480 x 480}.
The generated string is constrained to a \param{closed\_set}, similar to the formulation of classification tasks.

% \paragraph{Training:} 
% We finetune the pretrained model with cross-entropy optimization, similar to pretraining. 
% The batch size of $256$. 
% We use a cosine learning rate schedule with a peak rate of $5e-5$ and a number of steps of $40,000$.  

\subsubsection{Visual Grounding (RefCOCO)}

\paragraph{Task Introduction:}
Visual grounding requires the model to locate an image region based on a textual description. 
\ofasysstyled{} formulates this task as a sequence-to-sequence generation task.
The model takes a text and an image as input and generates a sequence of box tokens in an autoregressive manner.
We perform experiments on RefCOCO~\cite{refcoco,kazemzadeh14referitgame}. 
The standard metric Acc@\num{0.5} is reported on the corresponding validation set, that is, a bounding box is considered correct if $\text{IoU} \geq 0.5$ with the ground truth.

\paragraph{Default Instruction:} 
The default instruction for this task is as follows:
\begin{inst}
[IMAGE:img] which region does the text "[TEXT:cap]" describe? -> [BOX:region_coord]
\end{inst}
%During inference, we limit the maximum generated length to \num{4}.

% Where BOX denotes the continuous corner coordinates of the region bounding box. 
% The continuous coordinates are discretized to integers as bounding box tokens $\langle x1, y1, x2, y2 \rangle$ using the BOX preprocessor.
% \concept{add\_bos} denotes append $\langle bos \rangle$ to the begining of the bounding box tokens.
% \concept{add\_eos} denotes append $\langle eos \rangle$ to the end of the bounding box tokens.

% \paragraph{Training:} 
% For finetuning, the input image resolution is set to $512\times 512$. 
% We finetune the model on each dataset for \num{10} epochs with a batch size of \num{128}. 
% The learning rate is $3e-5$ with the label smoothing of \num{0.1}. 

\subsubsection{Grounded Image Captioning} 

\paragraph{Task Introduction:}
Grounded image captioning is the inverse task of visual grounding. 
Given an image and a region, the model is required to generate a description about the region. 
We use RefCOCO, RefCOCO+, RefCOCOg, and Visual Genome~\cite{visual_genome} as the pretraining datasets for this task.
This task is supposed to used in multi-task learning only and no inference support is currently included.

\paragraph{Default Instruction:} The default instruction for this task is as follows:
\begin{inst}
[IMAGE:img] what does the region describe? region [BOX:region_coord] -> [TEXT:cap]
\end{inst}

% \paragraph{Optimization:}
% This task is supposed to used in multi-task learning only and is not optimized for singe task performance by default.
% Although \ofasysstyled{} provides a set of trivial hyper-parameters, one should adjust the batch size, the learning rate, and the others accordingly.

\subsubsection{Object Detection}

\paragraph{Task Introduction:} 

Object detection is a common vision task that requires a model to recognize all objects in the image and localize their regions. 
We use OpenImages~\cite{openimages}, Object365~\cite{object365}, Visual Genome~\cite{visual_genome}, and COCO~\cite{chen15microsoft} as the pretraining datasets for this task.
This task is supposed to used in multi-task learning only and no inference support is currently included.

\paragraph{Default Instruction:} 

The default instruction for this task is as follows:
\begin{inst}
[IMAGE:img] what are the objects in the image? -> [ [BOX] [TEXT] ]*
\end{inst}
As the output is of variable lengths, a specific instruction parsing method is implemented via the \func{build\_instruction} method.
The \concept{[]*} notation is parsed to the actual number of the bounding boxes for a training example.

% \paragraph{Optimization:}
% Currently, this task is only considered in multi-task pretraining and is not optimized for singe task performance by default.
% Although \ofasysstyled{} provides a set of trivial hyper-parameters, one should adjust the batch size, the learning rate, and the others accordingly.

\subsubsection{Image Infilling}

\paragraph{Task Introduction:}
Image infilling task has been proved an effective task for both image and multi-modal pretraining ~\cite{mae,beit,wang22ofa}.
We mask the middle part of the raw images as input, and expect the model to restore the masked part from the corrupted input by generating the discrete codes produced by VQ-GAN models.
This self-supervised task is designed for multi-task pretraining only.

\paragraph{Default Instruction:} 
The default instruction for this task is as follows:
\begin{inst}
what is the complete image of "[IMAGE:img,mask_ratio=0.5]"? -> [IMAGE,preprocessor=image_vqgan,adapter=image_vqgan]
\end{inst}
% The text instruction is processed by the default text preprocessor and then the text adapter.
The input is the representation from the image pixels.
The input image resolution is set to \numproduct{256 x 256} and we mask the central \numproduct{128 x 128} part.
The attribute \param{mask\_ratio} is added to set the mask ratio of the image.
Following \cite{wang22ofa}, the target output is a seqeunce of discrete image codes generated by VQ-GAN~\cite{esser21taming}.
The output length in inference is fixed to \num{256} (\numproduct{16 x 16}), according to the image resolution and the compression ratio of VQGAN.
% The image are preprocessed by the VQ-GAN preprocessor.
% The VQ-GAN adapter, which is similar to text adapter but contains a 2D attention bias similar to the image, is then applied to the discrete token sequences.
% The post-processor maps the generated tokens back to image format.

% \paragraph{Training:} 
% This self-supervised task is designed for multi-task pretraining only.
% % During pretraining, the model is trained with batch size of \num{256} (for this task). 
% Although \ofasysstyled{} provides a set of trivial hyper-parameters, one should adjust the batch size, the learning rate, and the others accordingly.

\subsubsection{Image Generation (COCO Captions)}

\paragraph{Task Introduction:}

Text-to-Image generation has become a task that has attracted more and more attention of researchers as it demonstrates the excellent creation of neural network models \cite{dalle, cogview, latentdiffusion}. 
Similar to Image Infilling task, we use a VQ-GAN model to convert images into discrete codes, so that the sequence generator can generate a complete image by generating the code sequence autoregressively. 
We train our model on the train split of the MS COCO dataset and evaluate our model on the test split by randomly sampling \num{30000} images.
As for evaluation, following previous works \cite{wang22ofa, nvwa, huang2021unifying},  we use CLIP Similarity Score (CLIPSIM) to evaluate the semantic similarity between the query text and the generated images.

\paragraph{Default Instruction:} 

The default instruction for this task is as follows:
\begin{inst}
what is the complete image? caption: "[TEXT:cap]"? -> [IMAGE,preprocessor=image_vqgan,adapter=image_vqgan]
\end{inst}
We use the similar instruction and configuration like image infilling task to define the image generation task.
The main difference is that we use text instead of masked images as the input.
In inference, the output image resolution is set to \numproduct{256 x 256}, so the output length is  \num{1024} (\numproduct{32 x 32}) with respect to the compression ratio of VQ-GAN.

% \paragraph{Optimization:}
% For finetuning, the model is trained with a batch size of \num{512} and a learning rate of \num{1e{-3}} for \num{40000} step.
% During the evaluation, we sample \num{24} images and choose the best one to compute the metrics.
% The output image resolution is set to \numproduct{256 x 256}, so the output length is  \num{1024} (\numproduct{32 x 32}).

\subsection{Video-Related Tasks}

\subsubsection{Video Classification (Kinetics-400)}

\paragraph{Task Introduction:}
The video classification task is a fundamental task in the field of video understanding where the model needs to predict the label for a given video clip. 
We evaluate our model on the Kinetics-400 dataset~\cite{kay2017kinetics}, which contains \numk300k video clips from \num{400} classes. 
We report the accuracy on the val split of the Kinetics-400 dataset.

\paragraph{Default Instruction:} 
The default instruction for this task is as follows:
\begin{inst}
[VIDEO:video] what does the video describe? -> [TEXT:label_name,closed_set]
\end{inst}
We follow MViT~\cite{fan2021multiscale} for video data augmentation, which is incorporated into the \texttt{VIDEO} preprocessor. %during finetuning.

% \paragraph{Optimization:}
% We finetune the model using cross-entropy loss for \numk60k steps with a batch size of \num{512} and a learning rate of \num{5e{-5}}. The stochastic depth rate for encoder and decoder is set to $0.2$ and label smoothing is set to \num{0.1}. 

%\subsubsection{Video Generation}
%
%\paragraph{Task Introduction:}
%
%\paragraph{Default Instruction:} The default instruction for this task is as follows:
%
%
% TODO: IMAGE:video is the last frame of video.
%\begin{inst}
%<BOS> what is the next frame of the video? <EOS> category: [TEXT:label_name] video: [VIDEO:video] -> [IMAGE:video,preprocess=image_vqgan,adapter=image_vqgan,add_bos,add_eos]
%\end{inst}
%
%\paragraph{Training:} 

\subsubsection{Video Captioning (MSR-VTT)}

\paragraph{Task Introduction:}
The video captioning task requires the model to generate a textual description for a given video clip. 
We evaluate the proposed method on MSR-VTT caption dataset~\cite{xu2016msr}, which contains \numk10k video clips \numk200k descriptions of the videos. 
Following \cite{lin2022swinbert}, we report CIDEr~\cite{cider} scores on the val split of the MSR-VTT dataset.

\paragraph{Default Instruction:} 
The default instruction for this task is as follows:
\begin{inst}
[VIDEO:video] what does the video describe? -> [TEXT:cap]
\end{inst}
All data augmentation for videos are disabled except for the random flip. 

% \paragraph{Training:} 
% We finetune the model using cross-entropy loss for \numk1.6k steps with a batch size of \num{256} and a learning rate of \num{5e{-5}}. 
% The stochastic depth rate for encoder and decoder is set to $0.2$ and label smoothing is set to \num{0.1}. 

\subsubsection{Video Question Answering (MSR-VTT QA)}

\paragraph{Task Introduction:}
The video captioning task requires the model to generate a answer for a given video clip and a question related to that video clip. 
We evaluate the proposed method on MSR-VTT QA dataset~\cite{xu2017video}, which contains question-answer pairs extracted from the original MSR-VTT dataset~\cite{xu2016msr}.
We report the accuracy on the val split of MSR-VTT QA dataset.

\paragraph{Default Instruction:} 

The default instruction for this task is as follows:
\begin{inst}
[VIDEO:video] [TEXT:question] -> [TEXT:answer,is_label]
\end{inst}
We follow MViT ~\cite{fan2021multiscale} for video data augmentation. % during finetuning.

% \paragraph{Training:} 
% We finetune the model using cross-entropy loss for \numk6.4k steps with a batch size of \num{512} and a learning rate of \num{5e{-5}}. The stochastic depth rate for encoder and decoder is set to $0.2$ and label smoothing is set to \num{0.1}. 

\subsection{Audio-Related Tasks}

\subsubsection{Automatic Speech Recognition} 
% adapter: audio(wav、fbank)、phone
% loss: ctc、mam_loss、tacotron_loss
\paragraph{Task Introduction:}
% audio preprocessor & adapter
Automatic Speech Recognition (ASR) is the task of converting speech into sequences of discrete semantic tokens. 
We evaluate our model on the Librispeech~\cite{panayotov2015librispeech} and AISHELL-1~\cite{bu2017aishell} dataset. 
The Librispeech dataset contains \num{1000} hours of speech in English sampled at \num{16} kHz. 
The AISHELL-1 dataset contains \num{178} hours of Mandarin speech sampled at \num{16} kHz.
% OFAsys supports preprocessing speech features like waveform and mel-scale spectrogram, which are time-domain and frequency-domain features, respectively.
% Since speech sequences are much longer than text sequences, we introduce a CNN as the audio input adapter to down-sample the input speech signal.
% We consider a deeper CNN for waveform than that for Fbank to make the speech inputs have the same framerate. 
% The instruction for ASR is as following. Furthermore, the instruction can implement any speech-to-text tasks.

\paragraph{Default Instruction:} The default instruction for this task is as follows:
\begin{inst}
[AUDIO:wav] what is the text corresponding to the voice? -> [TEXT:text]
\end{inst}
To achieve better performance, the criterion for this tasks include an additional Connectionist Temporal Classification (CTC)~\cite{watanabe2017hybrid} loss alongside the sequence-to-sequence loss. 
Specifically, we input the encoder output matrix $X$ and the target sequence $Y$ for CTC loss, which computes ${P(Y|X)}$ by summing over the probability of all possible alignments between the two and maximizes the probability of ${P(Y|X)}$. 

% \paragraph{Training:} 
% For single task training, \ofasysstyled{} provides a default architecture the same as OFA-\textsc{Base} while the audio adapter contains a CNN with \num{768} channels, strides $(2,2)$ and kernel widths $(3,3)$, and \num{6} transformer layers.
% ctc_loss
% To achieve better performance, we train the ASR model with an additional Connectionist Temporal Classification (CTC)~\cite{watanabe2017hybrid} loss besides the sequence-to-sequence loss. 
% Specifically, we input the encoder output matrix $X$ and the target sequence $Y$ for the CTC module, which computes ${P(Y|X)}$ by summing over the probability of all possible alignments between the two and maximizes the probability of ${P(Y|X)}$. 
% The model is trained for \num{150000} steps with a batch size of \num{256}, and a learning rate of \num{1e{-3}}. The audio input is \num{80}-dimensional log Mel-filterbank features. 
% For finetuning based on OFA-\textsc{Base}, we train the model under the ASR and text-infilling tasks.

\subsubsection{Text-to-Speech} 

\paragraph{Task Introduction:}
% output是fbank特征，decoder prenet/postnet
Text-to-speech~(TTS) is the task of generating speech from input text.
We evaluate our model on the LJSpeech~\cite{ljspeech17} and BZNSYP\footnote{\url{https://www.data-baker.com/open_source.html}} datasets. 
The LJSpeech dataset contains \num{24} hours of English audio of a single speaker reading passages with a sample rate of \num{22050} Hz. 
The BZNSYP dataset includes \num{12} hours of Mandarin audio sampled at \num{48} kHz from a single speaker.
% Referring to Tacotron2, decoder inputs are first passed through the target audio input adapter implemented as a fully connected network. Furthermore, the decoder's transformer outputs are passed through the target audio output adapter implemented as a CNN, which can see the entire decoded sequence. Finally, the adapter outputs are added to the transformer outputs to improve the overall reconstruction. The instruction for TTS is:

\paragraph{Default Instruction:} 
The default instruction for this task is as follows:

\begin{inst}
[TEXT:text,preprocessor=text_to_phone] what is the voice corresponding to the text? -> [AUDIO:fbank,adapter=audio_tgt_fbank]
\end{inst}
\ofasysstyled{} converts original text into phonemes, then take phonemes as model input and output mel spectrograms. 
In inference, we use a well-trained vocoder HiFi-GAN~\cite{kong2020hifi} to transform the predicted mel spectrograms to the waveform.

% \paragraph{Training:} 
%  For single task training, the model is trained with a batch size of \num{256} and a learning rate of \num{1e{-03}} for \num{20000} steps.
% tts loss，stop token prediction
% Unlike text generation tasks, TTS outputs are continuous. Therefore, in this task, we use a mean squared error (MSE) loss rather than a cross-entropy loss. Besides, we also introduce a two-class classification subtask to predict whether the output sequence has been completed.

% \subsubsection{Speech-Text Joint Pretraining} 

% \paragraph{Task Introduction:}
% Recently, unsupervised pretraining has been shown be beneficial to learn good speech representation for downstream speech tasks. 
% It includes three sub tasks, that is, masked audio prediction speech to code and phone to text.

% \paragraph{Default Instruction:} The default instruction for this task is as follows:
% \begin{inst}
%     [AUDIO:wav] -> <BOS> [AUDIO_CODE:pseudo_code] <EOS>
% \end{inst}
% \begin{inst}
%     [PHONE:phone] -> <BOS> [TEXT:text] <EOS>
% \end{inst}

% \paragraph{Training:} 

\subsection{Structural Data--Related Tasks}

\subsubsection{Text-to-SQL (Spider)}

\paragraph{Task Introduction:}
Text-to-SQL could be considered as a semantic parsing task, which aims to generate executable SQL codes according to the question text and the information of corresponding database. 
In this task, model is supposed to not only truly understand the question text and database but also generate a SQL format code to solve the question.
We conduct our experiments on Spider dataset \cite{DBLP:conf/emnlp/YuZYYWLMLYRZR18}, which contain different complex SQL queries and different complex database in different domains. It consists of \num{10181} questions and \num{5693} unique complex SQL queries on \num{200} databases with multiple tables, covering \num{138} different domains. Some questions in the dataset is supposed to answer by cross-domain and cross-database semantic parsing problems.  
We use Exact Matching metric, measuring whether the generated SQL code as a whole is equivalent to the label SQL query. Following previous Text-to-SQL studies~\cite{DBLP:conf/emnlp/ZhongYK20}, we first decompose the SQL of both prediction and ground truth as bags of several components (SELECT, WHERE, GROUP BY, ORDER BY, KEYWORDS) and sub-components. 
The generated SQL code is correct only if all the components are correct compared with ground truth. 
The Exact Matching metric is the ratio of correct prediction among all the predictions.

\paragraph{Default Instruction:} 

The default instruction for this task is as follows:
\begin{inst}
[TEXT:src]; structured knowledge: "[STRUCT:database,preprocessor=database_to_text]". generating sql code. -> [TEXT:tgt] 
\end{inst}
Similar with \cite{DBLP:journals/corr/abs-2201-05966}, we consider the task as a sequence to sequence language task. \param{src} slot is the question text, and the \param{database} slot is the corresponding text format database information of the samples.
The database information is transformed by \func{table\_to\_text} into %``
% \func{$|$ [database name] $|$ [table\_1 name] : [column\_1 name] ( [mentioned row names] ), [column\_2 name], [column\_3 name], ... [table\_2 name] : ...}''
\begin{plaintext}
| [database name] | [table_1 name] : [column_1 name] ( [mentioned row names] ), [column_2 name], [column_3 name], ... [table_2 name] : ...
\end{plaintext}
The \param{mentioned row names} are the rows which are mentioned in the question. 
Adding the mentioned row names improves the performance in most cases~\cite{DBLP:journals/corr/abs-2201-05966}.

% \paragraph{Training:} 
% % ofasys中默认的训练相关优化参数。
% For single task training, \ofasysstyled{} provides a set of default hyper-parameters suited to the OFA-large checkpoint. The batch size is \num{64}, the learning rate is \num{5e{-5}}, the number of maximum updates is equivalent to \num{600} epochs, and the warmup steps are \qty{10}{\percent} of the total updates.
% For normalization in training, the dropout path of rate \num{0.1} is used. %The fp16 is closed. 
% The embedding of encoder and decoder are frozen. 

%  AS is shown in Table \ref{tab:spider_exact_match}, we compare all the models with similar model sizes. t5-base, bart-large and ofa-large.

% \begin{table}[h]
% \centering\caption{The Exact matching performancs of compared models in Spider.}
% %\vspace{-2mm}
% \begin{tabular}{|c|c|c|c|} 
% \hline
% Model & UnifiedSKG (t5-base) & Bart-large  & OFA-large \\
% \hline
%  & 58.10 & 48.26 & 62.90 \\
% \hline
% \end{tabular}\label{tab:spider_exact_match}
% \end{table}

\subsubsection{Table-to-Text (DART)}

\paragraph{Task Introduction:}
Table-to-Text \cite{DBLP:journals/jamia/CawseyWJ97,DBLP:conf/emnlp/LebretGA16} aims to describe a table by natural language. 
%It is a one-to-many task, since there are many ways to describe a table. 
%
We conduct experiments on DART \cite{DBLP:conf/naacl/NanRZRSHTVVKLIP21}, which is a triplet component table dataset. 
We consider the triplets as a three-column, multi-row table without column names. 
DART has \num{62659}, \num{5980}, and \num{12552} examples for training, validation, and testing, respectively.
The evaluation metric of Table-to-Text is BLEU~\cite{papineni02bleu} from SacreBLEU~\cite{post18call}. 
%we measure the BLEU score of prediction with the references.

\paragraph{Default Instruction:} 
The default instruction for this task is as follows:
\begin{inst}
structured knowledge: "[SRTUCT:database,preprocessor=table_to_text]". how to describe the tripleset? -> [TEXT:tgt]
\end{inst}
We consider the task as a sequence-to-sequence language task, where database is the table information following the format as %\func{[row1 col1] : [row1 col2] : [row1 col3] $|$ [row2 col1] : [row2 col2] : [row2 col3] $|$ ...}.
\begin{plaintext}
 [row1 col1] : [row1 col2] : [row1 col3] | [row2 col1] : [row2 col2] : [row2 col3] | ... 
\end{plaintext}

% \paragraph{Training:} 
% For single task training, \ofasysstyled{} provides a set of default hyper-parameters suited to the OFA-base checkpoint. The batch size is \num{192}, the learning rate is \num{5e{-5}}, the number of maximum updates is equivalent to 600 epochs, and the warmup steps are \qty{10}{\percent} of the total updates.
% For normalization in training, the dropout path of rate \num{0.1} is used. %The fp16 is closed. 
% The embedding of encoder and decoder are frozen. 
% \textbf{Performances.} In the Table \ref{tab:dart_bleu}, OFAsys outperforms not only other language model methods, but also current SOTA method HTML \cite{}.  
% \begin{table}[h]
% \centering\caption{The BLEU of compared methods in DART.}
% %\vspace{-2mm}
% \begin{tabular}{|c|c|c|c|c|} 
% \hline
% Model & UnifiedSKG (t5-base) & Bart-large & HTLM  & OFA-large \\
% \hline
%  & 46.22 & 45.50 & 47.10 & 47.89 \\
% \hline
% \end{tabular}\label{tab:dart_bleu}
% \end{table}

\subsubsection{TableQA (FeTaQA)}

\paragraph{Task Introduction:}
TableQA~\cite{DBLP:journals/corr/abs-2207-05270} is a question answering task according to a given table. 
We use FeTaQA dataset \cite{DBLP:journals/tacl/NanHMLVZKSKTMRT22} to evaluate our methods. 
FeTaQA is dataset based on \numk10k Wikipedia entries of (table, question, free-form answer, supporting table cells). 
We only use the table, question and free-form answers. 
The evaluation of TableQA is BLEU~\cite{papineni02bleu}.
%, we measure the BLEU score of prediction with the references.

\paragraph{Default Instruction:} 

The default instruction for this task is as follows:
\begin{inst}
structured knowledge: "[STRUCT:database,preprocessor=table_to_text]". what is the answer of the question "[TEXT:src]"? ->  [TEXT:tgt]
\end{inst}
The \param{src} is the question, the \param{database}  is the table, and the \param{tgt} is the predict answer.
The table format is the same as Table-to-Text.

% \paragraph{Training:} 
% For single task training, \ofasysstyled{} provides a set of default hyper-parameters suited to the OFA-large checkpoint. The batch size is \num{64}, the learning rate is \num{5e{-5}}, the number of maximum updates is equivalent to \num{600} epochs, and the warmup steps are \qty{10}{\percent} of the total updates.
% For normalization in training, the dropout path of rate \num{0.1} is used. The fp16 is closed. The embedding of encoder and decoder are frozen. 
% \textbf{Performances.} Among the compared methods in Table \ref{tab:FeTaQA_bleu}, t5-large is previous SOTA method. However, OFA-large achieves the best performance, even with less parameter size.

% \begin{table}[h]
% \centering\caption{The BLEU of compared methods in FeTaQA.}
% %\vspace{-2mm}
% \begin{tabular}{|c|c|c|c|} 
% \hline
% Model & UnifiedSKG (t5-base)  & t5-large  & OFA-large \\
% \hline
%  & 29.03 & 30.54 & 31.56 \\
% \hline
% \end{tabular}\label{tab:FeTaQA_bleu}
% \end{table}

\subsubsection{Sudoku}

\paragraph{Task Introduction:}
Sudoku is a common math puzzle game, which fills the blank of a \numproduct{9 x 9} tables with digits 1-9, such that each digit appears exactly once in each row, column, and $3 \times 3$ box.
%to let every such that every digit appears exactly once in each row, column and $3 \times 3$ box. 
Normally, a Sudoku has a single unique solution.
We use the Sudoku dataset in Kaggle\footnote{\url{https://www.kaggle.com/datasets/rohanrao/sudoku}}, which contains \numm10m puzzles, with difficulty from easy to hard.
The dataset is randomly split into \num{1000} samples for validation and \num{1000} for testing with the rest used for training.
We use Solved Acc as the evaluation metrics, which means the prediction that meets all the requirements is considered correct.

\paragraph{Default Instruction:} 

The default instruction for this task is as follows:
\begin{inst}
"[STRUCT:src,preprocessor=sudoku_to_text]". solve the sudoku. -> [STRUCT:tgt,preprocessor=sudoku_to_text]
\end{inst}
The \param{src} is the sudoku puzzles.
\func{sudoku\_to\_text} uses ``$:$'' to split columns and ``$|$'' to split rows into the form like %``\func{0 : 8 : 5 : 2 : 3 : 0 : 0 : 7 : 0 $|$ 1 : 4 : 0 : 8 : 0 : 9 : 0 : 0 : 0 $|$ 0 : 7 : 0 : 0 : 1 : 0 : 0 : 0 : 8 $|$ 7 : 0 : 9 : 0 : 0 : 5 : 0 : 0 : 3 $|$ 0 : 0 : 0 : 1 : 6 : 0 : 0 : 0 : 0 $|$ 5 : 0 : 2 : 3 : 0 : 0 : 0 : 1 : 0 $|$ 0 : 0 : 1 : 7 : 4 : 8 : 0 : 5 : 9 $|$ 6 : 5 : 0 : 9 : 0 : 3 : 0 : 0 : 0 $|$ 8 : 9 : 0 : 6 : 0 : 0 : 7 : 0 : 2}'',
\begin{plaintext}
 0 : 8 : 5 : 2 : 3 : 0 : 0 : 7 : 0 |
 1 : 4 : 0 : 8 : 0 : 9 : 0 : 0 : 0 |
 0 : 7 : 0 : 0 : 1 : 0 : 0 : 0 : 8 | 
 7 : 0 : 9 : 0 : 0 : 5 : 0 : 0 : 3 | 
 0 : 0 : 0 : 1 : 6 : 0 : 0 : 0 : 0 | 
 5 : 0 : 2 : 3 : 0 : 0 : 0 : 1 : 0 |
 0 : 0 : 1 : 7 : 4 : 8 : 0 : 5 : 9 | 
 6 : 5 : 0 : 9 : 0 : 3 : 0 : 0 : 0 | 
 8 : 9 : 0 : 6 : 0 : 0 : 7 : 0 : 2 |
\end{plaintext}
In the sequence, 0 means blank. %,  ``:'' split each digit and ``|'' split each line.
The \param{tgt} is the same format as \param{src},  replacing the 0 with answers.

% \paragraph{Training:} 
% For single task training, \ofasysstyled{} provides a set of default hyper-parameters suited to the OFA-large checkpoint. The batch size is \num{64}, the learning rate is \num{5e{-5}}, the number of maximum updates is equivalent to \num{50} epochs, and the warmup steps are \qty{10}{\percent} of the total updates.
% For normalization in training, the dropout path of rate \num{0.1} is used. %The fp16 is closed. 
% The embedding of encoder and decoder are frozen. 

% \textbf{Performances.} 
% The sudoku-net-v2 use 1M more puzzles than our model in Table \ref{tab:sudoku_acc}. OFA still get better result.
% \begin{table}[h]
% \centering\caption{The solved accuracy of sudoku.}
% %\vspace{-2mm}
% \begin{tabular}{|c|c|c|} 
% \hline
% Model & sudoku-net-v2  & OFA-large \\
% \hline
%  & 98.21 & 99.81 \\
% \hline
% \end{tabular}\label{tab:sudoku_acc}
% \end{table}

\subsection{Motion-Related Task}

\subsubsection{Text-to-Motion Synthesis}

\paragraph{Task Introduction:}
% OFASys can readily implement various motion synthesis tasks by merely providing an instruction and a dataset.
The text-to-motion synthesis task requires the model to generate a clip of human motion meeting the description of the given text. %, while other tasks such as music-to-dance and pose estimation (image/video-to-motion) can implemented almost in the same manner.
We use the text-motion pairs provided by KIT~\cite{Plappert2016}.
The AMASS~\cite{AMASS} dataset is also used to enrich the training data.
The text is set to an empty string when using AMASS, since AMASS does not provide text labels.
Unlike previous works~\cite{lin-vigil18,lang2pose}, we learn the rotation parameters rather than the positions of all the joints, in order to produce visually better results, which unfortunately means that we cannot directly compare our approach to the previous methods on the same benchmark.

\paragraph{Default Instruction:} 
The default instruction for the text-to-motion task is as follows:
\begin{inst}
motion capture: [TEXT:title] -> [MOTION:bvh_frames]
\end{inst}
The training criterion and the inference generator follow the DDPM method.
The accompanied preprocessor, adapter, and postprocessor are also available.
Please refer to \cref{appx:motion} for more details.

\section{More Experimental Settings}
\label{appx:settings}

%%%%%%%%
\begin{table}[t]
    \centering
    \small
    \setlength{\tabcolsep}{3pt}
    \begin{tabular}{@{}l l S S S@{}}
    \toprule
     \multirow{2}{*}{Task} &    \multirow{2}{*}{Dataset}   &  \multicolumn{2}{c}{{Specialist}} & \multicolumn{1}{c}{{Generalist \& Generalist MoE}} \\ \cmidrule(r){3-4} \cmidrule(l){5-5}
     & & {Batch Size} & {LR} & {Batch Size} \\ \midrule
     
     \multicolumn{5}{@{}l}{\textit{Text only tasks}} \\
     Instruction Tuning & NaturalInstruction v2 & 512 & 1e-5 & 512 \\
     Summarization & Gigaword  &  512 & 1e-4  & 512   \\\midrule
     \multicolumn{5}{@{}l}{\textit{Image tasks}}\\
     Classification & ILSVRC & 256 & 5e-5   & 2048 \\
     Visual Entailment & SNLI-VE  & 256 & 2e-5 & 256 \\
     Captioning  & COCO & 128 &  1e-5 & 2048 \\
     Visual Grounding  & RefCOCO & 128  & 3e-5  & 2048 \\
     Grounded Caption & RefCOCO & 256  & 1e-5    & 256 \\
     VQA & VQA v2 & 512 & 5e-5 & 1536 \\
     Image Generation & COCO & 512  & 1e-3  & 512 \\
     \midrule
     \multicolumn{5}{@{}l}{\textit{Audio tasks}} \\
     ASR & LibriSpeech & 256 & 1e-3  & 2048 \\
     TTS & LJSpeech  & 256 &  1e-3 & 1024 \\\midrule
     \multicolumn{5}{@{}l}{\textit{Video tasks}} \\
     Classification  & Kinetics400 &  512 & 5e-5  & 512 \\
     Captioning  & MSR-VTT   & 256 & 5e-5  & 128 \\
     VQA  & MSR-VTT QA & 512 &  5e-5  & 256
     \\\midrule
     \multicolumn{5}{@{}l}{\textit{Motion tasks}} \\
     Text-to-Motion & AMASS/KIT/AIST++ & 512 & 1e-3  & 2048 \\\midrule
     \multicolumn{5}{@{}l}{\textit{Other tasks}} \\
     Table-to-Text  & DART & 192 & 5e-5  & 128 \\
     Text-to-SQL  & Spider   &  64 &  5e-5 & 256 \\%\midrule
     % \multicolumn{5}{@{}l}{\textit{Zero-shot tasks}} \\
     % haha \\
     \bottomrule
    \end{tabular}
    \caption{Tasks, datasets, and main optimization hyper-parameters used in the experiments. We list the configurations of single-task and multi-task settings.}
    \label{tab:task parameter}
\end{table}
%%%%%%%%

The task mixture used in multi-task training include:
\begin{enumerate}[itemsep=0pt,parsep=0pt]
    \item Text summarization on Gigaword~\cite{gigaword}
    \item Instruction tuning on Natural-Instructions v2~\cite{natural_instruction_v2}
    \item Image classification on ImageNet-1K~\cite{imagenet}
    \item Image captioning on COCO Caption~\cite{chen15microsoft}
    \item Visual entailment on SNLI-VE~\cite{xie2019visual}
    \item Visual grounding on RefCOCO~\cite{refcoco}
    \item Grounded image captioning on RefCOCO~\cite{refcoco}
    \item Image generation on COCO Caption~\cite{chen15microsoft}
    \item Visual question answering on VQA v2~\cite{goyal17vqav2}
    \item Video classification on Kinetics400~\cite{imagenet}
    \item Video captioning on MSR-VTT~\cite{xu2016msr}
    \item Video question answering on MSR-VTT QA~\cite{xu2017video}
    \item Automatic speech recognition on LibriSpeech~\cite{panayotov2015librispeech}
    \item Text-to-speech on LJSpeech~\cite{ljspeech17}
    \item Table-to-text on DART~\cite{DBLP:conf/naacl/NanRZRSHTVVKLIP21}
    \item Text-to-SQL on Spider~\cite{DBLP:conf/emnlp/YuZYYWLMLYRZR18}
    \item Text-to-motion synthesis on AMASS~\cite{AMASS}, KIT~\cite{Plappert2016} and AIST++~\cite{li2021learn}

\end{enumerate}
\num{2} of these tasks, \ie, grounded image captioning and text-to-motion synthesis, does not come with validation sets and as a result, no scores are reported.

The optimization settings for those tasks are listed in \cref{tab:task parameter}.
For multi-task learning, as shown in \cref{sec:application}, a learning rate of 
\num{3e{-4}} is used. % and the maximum number of training updates is set to ??.
The batch size is altered to better suit the multi-task learning settings.
For single-task learning, each specialist is trained with a more appropriate set of hyper-parameters.

{\small
\bibliography{egbib}
\bibliographystyle{ieee_fullname}
}

\end{document}